\documentclass[lettersize, journal]{article}
\usepackage{csl}
\usepackage{amsmath,amsfonts,amssymb,mathtools}
\usepackage{bm}
\usepackage{graphicx}
\usepackage{array}
\usepackage{booktabs}
\usepackage{tabularx}
\usepackage{tabularray}
\UseTblrLibrary{booktabs}
\usepackage{multirow}
\usepackage{makecell}
\usepackage{adjustbox}
\usepackage{stfloats}
\usepackage{float}
\usepackage{placeins}
\usepackage{textcomp}
\usepackage{xurl}
\usepackage{url}
\usepackage{dsfont}
\usepackage{siunitx}
\usepackage[ruled,vlined,noend]{algorithm2e}
\usepackage{cite}
\usepackage[hidelinks]{hyperref}
\usepackage[nameinlink,capitalise,noabbrev]{cleveref}

\providecommand{\citep}[1]{\cite{#1}}

\begin{document}

\title{Scientific Machine Learning for Engine Health Management and Remaining Useful Life Prediction}

\cslauthor{Jostein Barry-Straume,\\Changmin Son, Adrian Sandu, \\Gavan Burke, Rekha Sundararajan, \\Andrew Rimell, James G. Steinrock}

\cslyear{26}
\cslreportnumber{3}
\cslemail{jostein@vt.edu}

\csltitlepage{}

\begin{abstract}
Engine Health Management (EHM) depends on reliable forecasting of Remaining Useful Life (RUL) and on tracking thermal indicators such as turbine gas temperature (TGT). In practice, real-world fleet data are heterogeneous and non-stationary, and point predictions alone are insufficient for risk-aware maintenance decisions. This paper presents a multi-task scientific machine learning framework for turbine prognostics that jointly predicts turbine gas temperature untrimmed (TGTU), Delta Turbine Gas Temperature (DTGT), and RUL, with quantified uncertainty in the form of prediction intervals whose empirical coverage is evaluated. A shared sequence encoder (convolutional front-end with residual bidirectional LSTM layers and attention pooling) feeds task-specific heads, including mean--variance estimation for probabilistic regression and, optionally, a survival head for threshold-based event modeling. The framework is designed to be tunable via a small set of practitioner-facing parameters (e.g., DTGT thresholding rules and RUL target construction) so that deployment can align with in-house policies and proprietary criteria. The predictive performance of the proposed framework is evaluated using both point and interval metrics, including mean absolute error (MAE), prediction interval coverage probability (PICP), mean prediction interval width (MPIW), and the coverage--width criterion (CWC). Results are reported both in aggregate and stratified by flight phase and maintenance segment to highlight operational-context effects and to support uncertainty-aware monitoring.
\end{abstract}

\section{Introduction}

Forecasting Remaining Useful Life (RUL) underpins maintenance strategies for turbofan gas engines. Accurate RUL forecasts have significant impacts in Engine Health Management (EHM), including: \textbf{(1) Safety}, by reducing the risk of catastrophic engine failure, \textbf{(2) Reliability}, by avoiding unplanned engine downtime, \textbf{(3) Cost-effectiveness}, by optimizing maintenance resources, and \textbf{(4) Lifespan extension}, by supporting proper monitoring of engine health.

Leveraging modern Scientific Machine Learning (SciML) approaches together with Uncertainty Quantification (UQ) provides a pathway to developing robust EHM frameworks. However, translating benchmark-level results into operational settings is challenging for at least two reasons. First, many published prognostic methods are developed and validated on simulated run-to-failure benchmarks (e.g., C--MAPSS), whereas real fleet data are heterogeneous, censored, and shaped by maintenance actions. Second, uncertainty estimates that appear adequate under i.i.d.\ assumptions can degrade under dataset shift, yet UQ is often omitted or not empirically validated in a way that supports risk-aware maintenance decisions.

In practice, real-world datasets contain missing values, sensor drift, varying mission and ambient conditions, and policy-driven maintenance resets that truncate trajectories. These factors require extensive data wrangling and preprocessing, and they motivate models that (i) align target definitions with in-house policy (e.g., what constitutes an ``event'' for RUL) and (ii) report uncertainty in a verifiable way through calibrated prediction intervals and reliability metrics. This emphasis on real data is important because it exposes non-stationarity and operational regime shifts that are typically absent in synthetic benchmarks and that can strongly affect both accuracy and calibration.

For this task, a deep learning (neural network) architecture for turbine prognostics is developed with the following core features: \textbf{(1) Multi-objective learning}. One unified network predicts turbine gas temperature untrimmed (TGTU), Delta Turbine Gas Temperature (DTGT), remaining useful life (RUL), and degradation/survival attributes from shared representations. \textbf{(2) Remaining useful life}. Multiple approaches to constructing and predicting RUL are supported within one system. \textbf{(3) Uncertainty Quantification}. Each target prediction is accompanied by prediction intervals with empirically evaluated coverage and sharpness. \textbf{(4) Tunable knobs}. Parameterized variables in the framework enable in-house tuning to adjust for proprietary information (e.g., DTGT thresholding rules and RUL target construction) that is not readily shareable with developers. \textbf{(5) Real-world evaluation}. The methodology is trained and evaluated on real fleet data and is analyzed by flight phase and maintenance segment to highlight operational-context effects.

The remainder of this paper is organized as follows. \Cref{ehm:sec:Background} reviews background information and common terminology used in EHM. \Cref{ehm:sec:Literature} surveys approaches to predicting RUL with machine learning methods. \Cref{ehm:sec:Methodology} describes the underlying methodologies that serve as the foundation of the present work. \Cref{ehm:sec:Framework} introduces the novel multi-objective Engine Health Management neural network. \Cref{ehm:sec:Setup} details the general experiment setup, data preprocessing, and neural network training. \Cref{ehm:sec:Results} presents the experimental results. \Cref{ehm:sec:Conclusions} summarizes the contributions of the paper and discusses potential directions for future work.

%%%%%%%%%%%%%%%%%%%%%%%%%%%%%%%%%%%%%%%%%%%%%%%

\section{Engine Health Management and Uncertainty Quantification}
\label{ehm:sec:Background}
This work studies aero-engine degradation monitoring and prediction with a focus on thermal indicators and predictive uncertainty. In this section, the following background concepts pertinent to the presented methodology are reviewed: Engine Health Management (EHM) in \cref{ehm:subsec:EHM}, Turbine Gas Temperature (TGT) in \cref{ehm:subsec:TGT}, Nominal Turbine Gas Temperature (NTGT) in \cref{ehm:subsec:NTGT}, Delta Turbine Gas Temperature (DTGT) in \cref{ehm:subsec:DTGT}, Remaining Useful Life (RUL) in \cref{ehm:subsec:RUL}, and Uncertainty Quantification (UQ) in \cref{ehm:subsec:UQ}.

\begin{figure}[H]
    \centering
    \includegraphics[width=\linewidth]{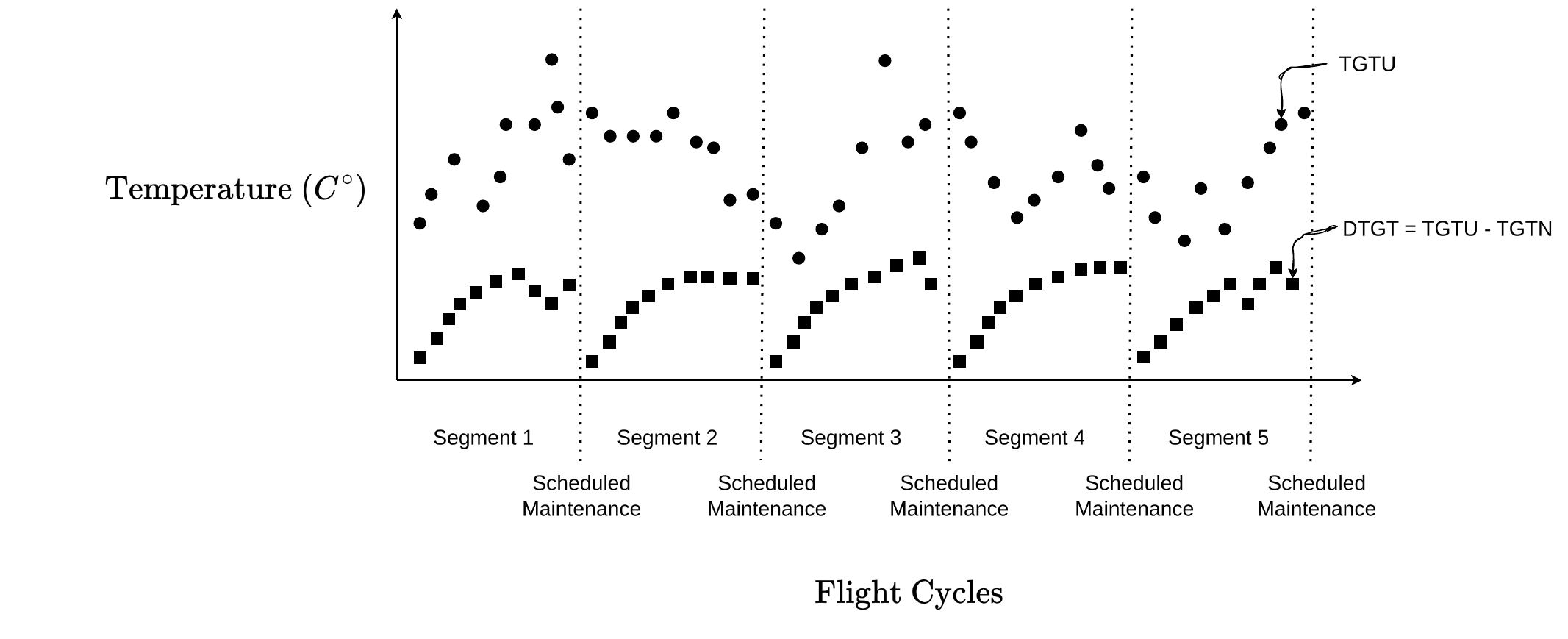}
    \caption{Degrading engine components require hotter operation to maintain thrust output. This is shown by the rise of DTGT over time within each maintenance segment.}
    \label{ehm:fig:tgt-dtgt-timeseries}
\end{figure}

Turbine Gas Temperature (TGT) critically affects component degradation and is therefore useful for predicting Remaining Useful Life (RUL). During the engine life cycle, operating temperatures degrade engine components, forcing the engine to operate at higher temperatures to maintain efficiency. This is illustrated in \cref{ehm:fig:tgt-dtgt-timeseries}, where the Delta Turbine Gas Temperature (DTGT) rises over time, but then resets after scheduled maintenance. DTGT is the difference between the recorded TGT and the analytical temperature from a physics-based model. This relationship between TGT, DTGT, and engine degradation can be leveraged to help better model Remaining Useful Life (RUL) for Engine Health Management (EHM).

\subsection{Engine Health Management (EHM)}
\label{ehm:subsec:EHM}
Engine Health Management (EHM), often framed within Prognostics and Health Management (PHM), integrates sensing, modeling, diagnostics, prognostics, and decision support to maintain safety and reduce life-cycle cost \cite{Kim2017PHM}. For gas turbines, EHM traditionally centers on gas-path measurements (temperatures, pressures, spool speeds, fuel flow) and compares them against baseline expectations to infer health states of major modules (compressor, combustor, turbine) \cite{Volponi2014EHMTrends,Fentaye2019GasPathReview}. Data-driven and model-based gas-path analysis are both used: the former leverages statistical/machine-learning surrogates; the latter adapts physics engine models and solves an inverse problem to estimate component health parameters \cite{Volponi2014EHMTrends,Fentaye2019GasPathReview,Tahan2017AppliedEnergyReview,Doel1994WLS}.

Model-based health estimation methods frequently use optimal state-estimation and filtering techniques (e.g., Kalman-filter variants) to infer latent health parameters from noisy sensor streams \cite{Simon2008Filtering}.

In EHM there are three main strategies: (1) Reactive maintenance, (2) Preventive maintenance, and (3) Predictive maintenance \cite{mathworks_three_ways_rul_ebook, ulusoy2018_predictive_maintenance_part1, mathworks_predictive_maintenance_overview, mathworks_rul_estimator_models, MathWorks2018OvercomingObstacles}. Reactive maintenance is the approach of running an engine to failure, and then conducting maintenance. Maintenance after failure is expensive and dangerous for the aerospace industry. Preventive maintenance is the approach of performing repair well before failure. This strategy typically does not consider the actual condition of the engine but mitigates risk at the expense of higher costs. Predictive maintenance seeks to optimize the time to do maintenance. In other words, this approach is the ``Goldilocks,'' in the sense that it is neither too early nor too late. By factoring in the condition of the engine, unplanned downtime and operational costs are minimized while providing anomaly detection.

\subsection{Turbine Gas Temperature (TGT)}
\label{ehm:subsec:TGT}
Turbine Gas Temperature (TGT) (closely related to the commonly trended exhaust gas temperature, EGT) is a primary thermal indicator of overall core loading and engine margin. Empirically, EGT/TGT correlates with operating condition (e.g., thrust, fuel flow, ambient temperature) and with gradual performance deterioration (e.g., compressor/turbine efficiency and flow capacity shifts) \cite{Yilmaz2009CFM56EGT,Fentaye2019GasPathReview}. Because raw temperature depends on flight and ambient conditions, effective monitoring requires normalizing against a baseline or ``nominal'' model before drawing health conclusions \cite{Fentaye2019GasPathReview,Doel1994WLS}.

\subsection{Nominal TGT (NTGT)}
\label{ehm:subsec:NTGT}
Nominal turbine gas temperature $\mathrm{NTGT}$ is the baseline (healthy) temperature predicted from observed controls/operating conditions (ambient state, Mach/altitude, shaft speeds, fuel flow, etc.). In practice, $\mathrm{NTGT}$ is generated either by (i) an adapted thermo-system model (gas-path analysis) \cite{Doel1994TEMPER,Doel1994WLS} or (ii) a learned baseline regressor trained on healthy fleet data \cite{Wang2023EGTBaseline}. Formally, for a feature vector $\mathbf{x}_t$ of operating and environmental variables, the baseline regressor has the form
\begin{equation}
\label{ehm:eq:nominal-tgt-baseline}
    \widehat{\mathrm{TGT}}_{N,t} \;=\; f(\mathbf{x}_t;\,\theta),
\end{equation}
where $f(\cdot)$ represents the baseline model (physics-based or data-driven) and $\theta$ its parameters. Baseline modeling is central to separating environmental/operational effects from health effects \cite{Fentaye2019GasPathReview,Doel1994WLS,Wang2023EGTBaseline}.

\subsection{Delta TGT (DTGT)}
\label{ehm:subsec:DTGT}
Given the measured $\mathrm{TGT}_t$ and its nominal counterpart $\widehat{\mathrm{TGT}}_{N,t}$, the residual is defined as
\begin{equation}
\label{ehm:eq:dtgt-residual}
    \mathrm{DTGT}_t \;=\; \mathrm{TGT}_t \;-\; \widehat{\mathrm{TGT}}_{N,t}.
\end{equation}
This ``delta'' is a standard gas-path residual (measured minus baseline) used to trend degradation, trigger alerts, and support fault isolation \cite{Doel1994WLS,Fentaye2019GasPathReview}. Because $\mathrm{DTGT}$ largely removes confounding from ambient and mission profiles, persistent positive drift is indicative of shrinking thermal margin and evolving loss mechanisms (e.g., compressor fouling or turbine efficiency loss) \cite{Fentaye2019GasPathReview,Yilmaz2009CFM56EGT}. Robust baseline estimation is therefore critical to high-fidelity $\mathrm{DTGT}$ modeling \cite{Wang2023EGTBaseline}.

\subsection{Remaining Useful Life (RUL)}
\label{ehm:subsec:RUL}
RUL is the time from now until an asset's health trajectory meets a failure or maintenance threshold. Canonical estimation approaches include model-based (state-space and physics-of-failure) and data-driven (statistical/ML) methods, whereas hybrid schemes combine both \cite{Heng2009PrognosticsReview,Si2011RULReview}. Reviews of the RUL literature detail degradation process modeling, covariate effects, censored operation, learning from fleets, and evaluation metrics \cite{Heng2009PrognosticsReview,Si2011RULReview}. Moreover, the ``best" approach depends on data availability, observability of degradation, and the decision context. In thermal-margin monitoring, $\mathrm{DTGT}$ (or analogous deltas) often serves as an interpretable health indicator that feeds the prognostic model.

\subsection{Uncertainty Quantification (UQ)}
\label{ehm:subsec:UQ}
Uncertainty quantification (UQ) refers to characterizing how confident a model is in its predictions---for example by outputting predictive distributions or prediction intervals whose empirical coverage can be evaluated. UQ is indispensable for EHM because maintenance decisions are risk-based. Two complementary views are prevalent. (1) In statistical learning, one quantifies predictive dispersion via probabilistic regression, ensembling, Bayesian methods, or prediction intervals; quality is judged by calibration and coverage/width trade-offs \cite{Khosravi2011Review,Abdar2021UQReview}. (2) In PHM, uncertainty sources (initial state, model form/parameters, future loads, sensor noise) are parsed and propagated to RUL distributions; both Bayesian and reliability-based viewpoints are used \cite{Sankararaman2015MSSP}. For deep models, aleatoric (data) and epistemic (model) uncertainty are commonly distinguished and jointly modeled \cite{KendallGal2018}. In this work, prediction intervals are reported for both health indicators (e.g., $\mathrm{DTGT}$) and RUL, enabling risk-aware decisions such as $\mathbb{P}(\mathrm{RUL}<\tau_{\mathrm{alarm}})$.

\begin{figure}[H]
    \centering
    \includegraphics[width=\linewidth]{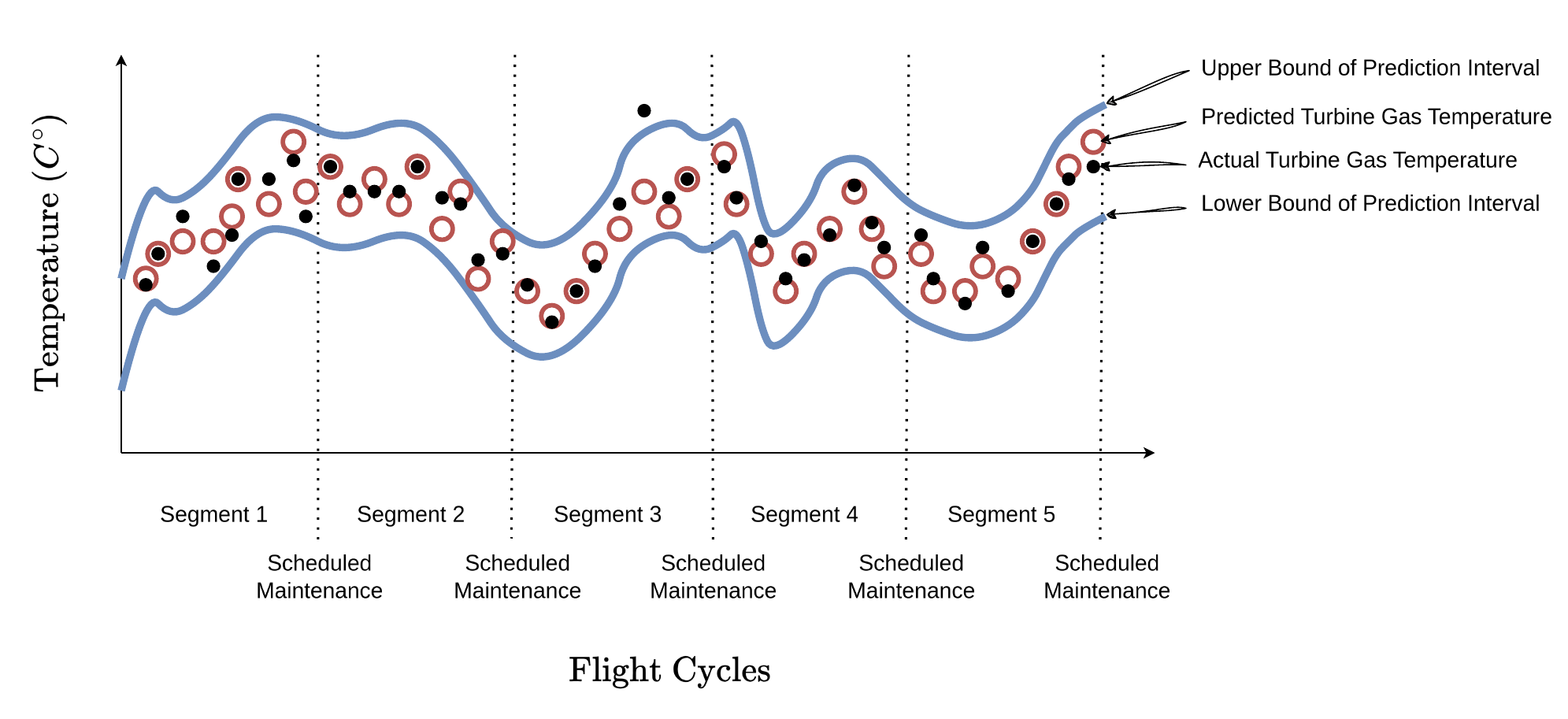}
    \caption{Illustration of prediction intervals for a time series of Turbine Gas Temperature (TGT) records versus predictions.}
    \label{ehm:fig:TGT-timeseries}
\end{figure}

\begin{figure}[H]
    \centering
    \includegraphics[width=0.35\linewidth]{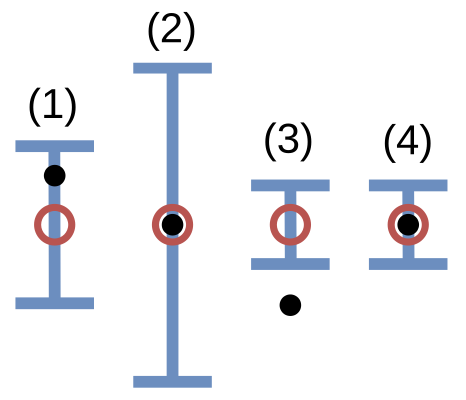}
    \caption{Illustration of different UQ scenarios. \textbf{(1.)} Valid coverage but inaccurate prediction. \textbf{(2.)} Valid coverage but unusable prediction width. \textbf{(3.)} Inaccurate prediction and invalid coverage. \textbf{(4.)} Accurate prediction and optimized valid coverage.}
    \label{ehm:fig:UQ-situations}
\end{figure}

EHM relies on comparing measured gas-path variables to trustworthy nominal baselines; the residual $\mathrm{DTGT}$ is a health-sensitive, operation-normalized indicator. RUL models translate such health indicators into time-to-threshold predictions, while principled UQ communicates confidence and supports decision-making. The proposed methods in this paper build directly on these foundations.

Recent studies show that nominally well-calibrated uncertainty estimates may deteriorate under dataset shift, motivating explicit evaluation across operating regimes \cite{Ovadia2019CanYouTrust}. In addition to coverage-based diagnostics, proper scoring rules provide principled evaluation of probabilistic forecasts \cite{GneitingRaftery2007Scoring}, and conformal prediction offers distribution-free prediction intervals under exchangeability assumptions \cite{AngelopoulosBates2021Conformal}.

%%%%%%%%%%%%%%%%%%%%%%%%%%%%%%%%%%%%%%%%%%%%%%%%%%%%%%%%%%%%%%%%%%%%%%

%%%%%%%%%%%%%%%%%%%%%%%%%%%%%
% Literature Review Section %
%%%%%%%%%%%%%%%%%%%%%%%%%%%%%
\section{Data-Driven RUL Prognostics for Turbofan Engines}
\label{ehm:sec:Literature}

\paragraph{Scope and Dataset Conventions}
Most deep-learning approaches for predicting remaining useful life (RUL) of aero engines are developed and evaluated on NASA's turbofan simulation datasets \cite{saxena2008nasa}. The dataset terminology in this paper follows that of \cite{saxena2008nasa} while avoiding shorthand split codes: when necessary, it denotes ``single-operating-condition'' or ``multi-operating-condition'' subsets of C--MAPSS, and ``single-fault'' versus ``multi-fault'' settings. \emph{N--CMAPSS} denotes the higher-fidelity benchmark dataset.

%%%%%%%%%%%%%%%%
% CNN          %
%%%%%%%%%%%%%%%%

\subsection{Convolutional Neural Network (CNN)}

In \cite{Giduthuri2016Deep}, Giduthuri et~al. introduce one of the first CNN regressors for multivariate engine time series, applying one-dimensional convolutions along time to learn features directly from raw sensors. A strength of the approach is its end-to-end feature learning for RUL without hand-crafted indicators. However, the approach offers no pathway for quantifying uncertainty. Moreover, success is demonstrated only on simulated C--MAPSS data.

A widely used deep-learning baseline in this category is the convolutional approach of Li et~al. \cite{Li2018DCNNRUL}, which demonstrates strong point-estimation performance on C--MAPSS but does not explicitly optimize calibrated prediction interval coverage.

In \cite{DeVol2021InceptionRUL}, DeVol et~al. adapt Inception CNNs to one-dimensional time series and validate their success on the N--CMAPSS benchmark. A strength of the approach is achieving competitive accuracy on high-fidelity data using parallel multi-scale temporal filters. A limitation of this approach is the lack of model-based explainability. Additionally, while the framework is competitively accurate, its predictions do not include quantified uncertainty.

In \cite{SolisMartin2021Stacked}, Solis-Martin et~al. leverage two stacked CNNs for RUL prediction. The first CNN extracts features from normalized raw data streams, while the second CNN performs RUL regression. Strengths of this approach include incorporating $k$-fold cross validation for model parameter selection, with Bayesian optimization enabling confidence intervals for RUL predictions. The framework's lower performance on the hidden validation set of C--MAPSS compared with the cross-validation score indicates possible overfitting on the training dataset.

%%%%%%%%%%%%%%%%
% LSTM         %
%%%%%%%%%%%%%%%%
\subsection{Long Short-Term Memory (LSTM) and CNN Hybrids}

In \cite{Hong2020RUL}, Hong et~al. combine CNN, LSTM, and bi-directional LSTM layers into one framework for RUL prediction on the C--MAPSS dataset. A strength of the approach is leveraging Shapley additive explanation (SHAP) techniques to identify input variables that have a significant impact on prediction results. Although the approach offers post-hoc explainability, it does not provide any quantified measures of uncertainty.

In \cite{Peng2021RULTurbofan}, Peng et~al. use CNN and LSTM in parallel fashion to provide RUL prediction on the C--MAPSS single- and multi-condition datasets. Data from C--MAPSS are preprocessed, then used as input variables for both a CNN and an LSTM. The subsequent outputs of these networks are then passed into a deep fully connected CNN for RUL regression. A strength of this approach is its use of the CNN branch of the architecture for spatial input data, and likewise the LSTM branch for temporal input data. Uncertainty quantification is not addressed in the paper.

In \cite{Li2022CNNLSTMAttention}, Li et~al. improve upon the CNN and LSTM hybrid architecture by incorporating a convolutional block attention module (CBAM). This CBAM module is placed between the CNN and LSTM to identify key variables related to RUL. Results on the C--MAPSS data show an improvement in accuracy compared to prior approaches. There is no formal prediction interval coverage or reliability assessment.

%%%%%%%%%%%%%%%%%%%%%%%
% Attention Mechanism %
%%%%%%%%%%%%%%%%%%%%%%%
\subsection{Attention Mechanism}

In \cite{daCosta2019AttentionLSTM}, Da Costa et~al. propose a global-attention LSTM for turbofan degradation with interpretable time-step weights. A strength of this approach is its ability to yield competitive RUL prediction without requiring previous degradation information. Visual inspection of the attention weights offers insight into the architecture's focus on input sensor data throughout the lookback window of input data.

In \cite{Tian2023SpatialTemporalLSTM}, Tian et~al. introduce a spatial-correlation and temporal-attention LSTM that models inter-sensor relations jointly. A strength of this approach is its stronger inductive bias for multivariate coupling. However, attention weights are not a guarantee of feature importance, and correlation does not imply causation. Success is demonstrated via the C--MAPSS dataset.

In \cite{Qin2022TemporalDeepDegradation}, Qin et~al. put forth their Temporal Deep Degradation Network (TDDN) to learn attention-mechanism-enhanced latent degradations over time. The TDDN is a neural network with a one-dimensional CNN front-end, fully connected layers in the middle, and an attention layer on the back end. Attention weights are visualized through the life cycle of the engine, ranging from healthy to failure, highlighting how the model focuses on the moving window of input data as the engine deteriorates. This paper does not treat uncertainty quantification or prediction interval coverage.

In \cite{Wang2025DualAttention}, Wang et~al. combine CNN-Channel Attention Mechanism (CNN-CAM) and Gated Recurrent Unit Self-Attention Mechanism (GRU-SAM) for a dual attention approach to RUL prediction. Spatial features are extracted via the CNN-CAM, whereas temporal features are extracted from GRU-SAM. A strength of this approach is its ability to improve prediction accuracy by dynamically assigning different weights to both spatial and temporal features. Success is demonstrated through the C--MAPSS dataset, albeit without quantified uncertainty.

%%%%%%%%%%%%%%%%
% Autoencoders %
%%%%%%%%%%%%%%%%
\subsection{Autoencoders}

In \cite{Liu2023SAETCN}, Liu et~al. present a Sparse Autoencoder Temporal Convolutional Network (SAE-TCN) to predict RUL in the C--MAPSS dataset. A strength of this approach is using the autoencoder to denoise data and extract features, while the TCN part of the network trains on low-dimensional data provided by the autoencoder. Accuracy metrics are provided for pointwise prediction without uncertainty intervals and without coverage metrics.

In \cite{Fan2024TwoStageTransformer}, Fan et~al. introduce their two-stage attention-based hierarchical transformer (STAR) framework. Their STAR methodology is a hierarchical encoder--decoder with a two-stage attention block that first captures temporal relations, then sensor-wise relations, alongside multi-scale (patch-merging) temporal abstraction. Success is demonstrated via comprehensive experimental results on all standard C--MAPSS datasets. The STAR architecture does not account for uncertainty quantification in its predictions. This line of work ultimately traces to the transformer self-attention architecture introduced by Vaswani et~al. \cite{Vaswani2017Attention}.

%%%%%%%%%%%%%%%%%%%%%%%%%%%%%%
% Trends of Prior Approaches %
%%%%%%%%%%%%%%%%%%%%%%%%%%%%%%
\subsection{Trends of Prior Approaches}
A clear trajectory in the literature is the shift from local, short-horizon temporal feature extraction toward architectures that can represent long-range degradation dynamics. Early work often relied on 1D CNNs applied along the time axis \cite{Giduthuri2016Deep}, which can effectively capture local temporal motifs but are limited in their effective receptive field. These approaches progressively gave way to CNN--LSTM hybrids and attention-augmented recurrent models \cite{Hong2020RUL,Li2022CNNLSTMAttention,daCosta2019AttentionLSTM,Tian2023SpatialTemporalLSTM}, as well as temporal convolutional networks (TCNs) designed to extend temporal context without recurrence \cite{Liu2023SAETCN}. More recently, hierarchical Transformer designs that explicitly separate temporal attention from sensor-wise attention have emerged \cite{Fan2024TwoStageTransformer}. Overall, the dominant trend is toward systematically increasing temporal receptive field and representational capacity so that models can leverage longer degradation trajectories rather than only short-term correlations.

At the same time, sensor-wise structure has become increasingly explicit in model design. Rather than treating multivariate sensor streams as an unstructured feature vector, many works now incorporate a notion of ``spatial'' (sensor) structure through dedicated CNN fusion pathways or sensor-wise attention mechanisms \cite{Peng2021RULTurbofan,Li2022CNNLSTMAttention,Fan2024TwoStageTransformer}. This architectural choice reflects the reality that turbofan sensors are correlated and that operating-context variability can induce structured patterns across channels. In contrast, explainability is typically introduced as an add-on rather than a native design principle: SHAP and feature-selection pipelines are commonly applied post hoc \cite{Hong2020RUL}, and while attention maps can highlight salient time steps, they do not guarantee causal importance \cite{daCosta2019AttentionLSTM,Tian2023SpatialTemporalLSTM}. Empirically, the field remains strongly benchmark-driven, with most studies relying on C--MAPSS and a smaller subset evaluating on the newer N--CMAPSS with more realistic flight profiles \cite{DeVol2021InceptionRUL}; as a result, robustness to non-stationary operating regimes is often less explored than tuning for benchmark performance. Finally, uncertainty is rarely calibrated in a way that supports decision-making: most approaches do not center uncertainty quantification (UQ) objectives, and coverage guarantees, interval calibration, and reliability under distribution shift are typically not reported \cite{Ovadia2019CanYouTrust}.

%%%%%%%%%%%%%%%%%%%%%%%%%%
% Gaps and Opportunities %
%%%%%%%%%%%%%%%%%%%%%%%%%%
\subsection{Gaps and Opportunities}
One major opportunity is to elevate calibrated uncertainty and reliability to first-class objectives in turbofan prognostics. While contemporary models often emphasize point accuracy, they seldom deliver calibrated prediction intervals or reliability metrics under distribution shift, leaving a gap between strong benchmark results and actionable maintenance decision support. This motivates integrating mean--variance estimation (MVE), conformal calibration, deep ensembles, and proper interval scoring into turbofan RUL workflows, alongside explicit reporting of coverage and width behavior under both in-distribution and shifted conditions.

Another opportunity is to expand turbofan prognostics beyond single-task RUL prediction to multi-task learning over complementary, physically meaningful targets---for example turbine gas temperature indicators (TGT/DTGT), degradation rates, and RUL---using shared backbones that can regularize representations and yield interpretable, engineering-relevant outputs. In addition, generalization across operating regimes remains under-tested: although multi-operating-condition subsets exist, systematic cross-regime validation and explicit domain-shift studies (e.g., covariate or conditional shift) are still uncommon, suggesting room for methods that target domain robustness or meta-learning. There is also a gap in physics- and policy-awareness, since monotone RUL trajectories, maintenance thresholds, and cost-aware decision constraints are rarely encoded; architectures that enforce or post-process monotonicity and explicitly connect predictions to maintenance-policy objectives could improve trustworthiness. Finally, interpretability methods could be strengthened beyond attention and post-hoc SHAP/feature selection \cite{Hong2020RUL}, for example through task-specific sensitivity analyses and counterfactual probes that more directly support engineering rationale and hypothesis testing.

\begin{table*}[!t]
\centering
\small
\caption{Literature review overview.}
\label{ehm:table:lit-review}
\renewcommand{\arraystretch}{1.1}
%\begin{adjustbox}{width=\textwidth,totalheight=\textheight,keepaspectratio}
%\resizebox{\linewidth}{!}{%
%\begin{tabularx}{\textwidth}{l X X X X}

\begin{tabularx}{\textwidth}{>{\setlength\hsize{.15\hsize}} l >{\setlength\hsize{.15\hsize}} X >{\setlength\hsize{.15\hsize}} X >{\setlength\hsize{.25\hsize}} X >{\setlength\hsize{.30\hsize}} X}

\toprule
\textbf{Paper} & \textbf{Model family} & \textbf{Data} & \textbf{Key idea/feature} & \textbf{Typical limitations} \\
\midrule
\cite{Giduthuri2016Deep} & CNN (1D temporal) & C--MAPSS (sim) & End-to-end temporal feature extraction from raw sensors & No UQ; simulation-only; limited regime analysis \\
\cite{DeVol2021InceptionRUL} & Inception CNN (1D) & N--CMAPSS & Multi-scale temporal filters; strong challenge results & No explainability/UQ; point estimates \\
\cite{SolisMartin2021Stacked} & Stacked DCNN & Challenge data (turbofan) & Two-level CNN: extractor then regressor; Bayesian Optimization for selection & Single-task; calibration not addressed \\
\cite{Hong2020RUL} & CNN+LSTM (xAI) & C--MAPSS & Dimensionality reduction + Shapley additive explanation (SHAP) explanations & Post-hoc xAI; no interval coverage \\
\cite{Peng2021RULTurbofan} & LSTM + FCN fusion & C--MAPSS & Temporal (LSTM) + sensor-wise (CNN) feature fusion & Heuristic fusion; no UQ \\
\cite{daCosta2019AttentionLSTM} & Attention LSTM & C--MAPSS & Global temporal attention for interpretability & Attention $\neq$ causality; calibration absent \\
\cite{Tian2023SpatialTemporalLSTM} & Spatial-temporal LSTM & C--MAPSS & Inter-sensor correlation + temporal attention & Complexity; no reliability guarantees \\
\cite{Li2022CNNLSTMAttention} & CNN--LSTM + CBAM & C--MAPSS & Channel \& spatial attention before LSTM & Point-only; no shift analysis \\
\cite{Liu2023SAETCN} & SAE--TCN & C--MAPSS & Denoising + long-range temporal convolutions & Lacks calibrated UQ; transfer underexplored \\
\cite{Qin2022TemporalDeepDegradation} & Degradation-attention net & C--MAPSS & Learns latent degradation dynamics with attention & Assumptions implicit; no calibration \\
\cite{Fan2024TwoStageTransformer} & Hierarchical Transformer & C--MAPSS & Two-stage (temporal then sensor-wise) attention; multi-scale & Heavy; intervals/coverage not reported \\
\cite{Wang2025DualAttention} & CNN+GRU dual attention & C--MAPSS & Channel attention (spatial) + self-attention (temporal) & General-purpose; turbofan-specific shift not deep-dived \\
\bottomrule
\end{tabularx}
%\end{adjustbox}
%}
\end{table*}

%%%%%%%%%%%%%%%%%%%%%%%
% Literature Overview %
%%%%%%%%%%%%%%%%%%%%%%%
\subsection{Literature Synthesis}
Across a decade, RUL research progressed from temporal CNNs to hybrid CNN--RNNs, to TCNs and Transformers. The consistent gains come from (i) enlarging temporal context, (ii) modeling sensor-wise structure explicitly, and (iii) modest use of post-hoc interpretability. However, the literature rarely addresses calibrated uncertainty, policy-aware monotonicity, and cross-regime robustness. The presented framework in this paper answers these gaps with a multi-task, attention-equipped temporal backbone whose outputs (network heads) produce both point predictions and calibrated uncertainties, evaluated under operating-regime shifts with engineering-aligned metrics and constraints. \Cref{ehm:table:lit-review} provides a tabular summary of the prior approaches discussed in this literature review.

\paragraph{Positioning relative to foundational deep-learning and UQ literature}
Foundational deep-learning approaches such as deep convolutional networks for RUL prediction \cite{Li2018DCNNRUL} and self-attention transformer sequence models \cite{Vaswani2017Attention} motivate many of the architectures in \cref{ehm:table:lit-review}. While these models can achieve strong point-estimation accuracy on C--MAPSS, they are often evaluated primarily under i.i.d.\ assumptions, and calibrated uncertainty under dataset shift is less frequently examined. Empirical evidence shows that common uncertainty estimators can fail under dataset shift \cite{Ovadia2019CanYouTrust}, underscoring the need for calibration-aware evaluation. Beyond coverage-based diagnostics, proper scoring rules provide a principled basis for evaluating probabilistic forecasts \cite{GneitingRaftery2007Scoring}. Distribution-free conformal prediction methods offer complementary coverage guarantees under exchangeability assumptions, but can trade off sharpness for validity \cite{AngelopoulosBates2021Conformal}.

\subsection{Strengths, limitations, and novelty of the present work}
The proposed framework contributes a unified multi-head SciML model that couples data-driven feature learning with task-aligned probabilistic objectives (MVE regression, exponential degradation, and Weibull survival). A key novelty is the inclusion of a differentiable coverage--width regularizer in the training loss, enabling direct optimization of interval reliability rather than purely post hoc reporting. The paper also emphasizes phase/segment-level analysis of UQ and anomaly detection to better reflect operational interpretability. Limitations include reliance on non-run-to-failure real-world proprietary data and parametric Gaussian interval assumptions for the MVE heads; future work should validate on additional fleets and consider non-Gaussian or distribution-free interval constructions.

%%%%%%%%%%%%%%%%%%%%%%%%%%%%%%%%%%%%%%%%%%%%%%%%%%%%%%%%%%%%%%%%%%%%%%

\section{Methodology}
\label{ehm:sec:Methodology}

% ====================== Generic formulations (come first) ======================
\subsection{One-dimensional Convolution with Residual Blocks and Max Pooling}
Let $\mathbf{X}_{t-L+1:t}\in\mathbb{R}^{L\times d}$ denote the lookback window of length $L$ and feature dimension $d$ ending at time $t$.
A residual temporal block applies two 1D convolutions and a skip connection:
\begin{align}
  \mathbf{U}^{(\ell)} &= \sigma\!\Big(\mathrm{BN}\!\big(W^{(\ell)}_2 * \sigma\!\big(\mathrm{BN}(W^{(\ell)}_1 * \mathbf{Z}^{(\ell)})\big)\big)\Big) + \mathbf{Z}^{(\ell)}, \label{ehm:eq:gen-cnn-resblock}
\end{align}
\emph{where} ${*}$ is convolution along the time axis, $W^{(\ell)}_1,W^{(\ell)}_2$ are convolution kernels at block $\ell$, $\mathrm{BN}(\cdot)$ is batch normalization, $\sigma(\cdot)$ is a pointwise nonlinearity (e.g., ReLU), $\mathbf{Z}^{(\ell)}$ is the block input of shape $\mathbb{R}^{T^{(\ell)}\times w^{(\ell)}}$, and $\mathbf{U}^{(\ell)}$ has the same shape.
Max pooling with stride $s_{\mathrm{pool}}$ reduces length via $T^{(\ell+1)}=\big\lfloor T^{(\ell)}/s_{\mathrm{pool}}\big\rfloor$, and $\mathbf{Z}^{(\ell+1)}$ is typically a linear map of $\mathbf{U}^{(\ell)}$ to the desired channel width.
After $D$ blocks with $s_{\mathrm{pool}}=2$,
\begin{align}
  T' &= \big\lfloor L/2^{D}\big\rfloor, \label{ehm:eq:gen-cnn-length}
\end{align}
\emph{where} $T'$ is the post-CNN sequence length passed to the recurrent backbone. Residual CNNs follow \cite{He2016ResNet}.

\subsection{Residual Bi-directional Long Short-Term Memory}
Given CNN features $\mathbf{Z}\in\mathbb{R}^{T'\times w}$, a bidirectional LSTM layer produces forward/backward states $\overrightarrow{\mathbf{h}}_i,\overleftarrow{\mathbf{h}}_i$ for $i=1,\dots,T'$, and outputs
\begin{align}
  \mathbf{h}_i &= \big[\overrightarrow{\mathbf{h}}_i;\ \overleftarrow{\mathbf{h}}_i\big]\in\mathbb{R}^{h}. \label{ehm:eq:gen-bilstm-one}
\end{align}
Stacking with residual connections yields
\begin{align}
  \mathbf{H}^{(0)} &= \mathbf{Z}, \quad
  \mathbf{U}^{(\ell)} = \mathrm{BiLSTM}^{(\ell)}\!\big(\mathbf{H}^{(\ell-1)}\big), \quad\\
  \mathbf{H}^{(\ell)} &= \mathbf{U}^{(\ell)} + \mathbf{H}^{(\ell-1)}, \ \ell=1,\dots,L_{\mathrm{rnn}}, \label{ehm:eq:gen-bilstm-stack}
\end{align}
\emph{where} $\mathrm{BiLSTM}^{(\ell)}(\cdot)$ is the $\ell$-th bidirectional LSTM with hidden size $h$, $\mathbf{H}^{(\ell)}\in\mathbb{R}^{T'\times h}$ is the residual output, and $L_{\mathrm{rnn}}$ is the number of stacked layers \cite{Hochreiter1997,SchusterPaliwal1997}.

\subsection{Attention Pooling}
A single-head additive attention produces a fixed-size summary from $\mathbf{H}=\mathbf{H}^{(L_{\mathrm{rnn}})}$:
\begin{align}
  e_i &= \mathbf{v}^{\top}\tanh(W \mathbf{h}_i + \mathbf{b}), \nonumber\\
  \alpha_i &= \frac{\exp(e_i)}{\sum_{j=1}^{T'} \exp(e_j)}, \nonumber\\
  \mathbf{c} &= \sum_{i=1}^{T'} \alpha_i\,\mathbf{h}_i. \label{ehm:eq:gen-attn}
\end{align}
\emph{where} $W\in\mathbb{R}^{a\times h}$, $\mathbf{b}\in\mathbb{R}^{a}$, and $\mathbf{v}\in\mathbb{R}^{a}$ are trainable attention parameters, $\alpha_i\!\in\!(0,1)$ are weights that sum to one, and $\mathbf{c}\in\mathbb{R}^{h}$ is the context vector used by downstream heads \cite{Bahdanau2015}.

\subsection{Mean--Variance Estimation (MVE) using the Gaussian Negative Log-Likelihood Loss Function}
For any scalar target $y$, the head outputs $(\mu,\sigma)$ and models
\begin{align}
  y \mid \mathbf{c} &\sim \mathcal{N}\!\big(\mu(\mathbf{c}),\,\sigma(\mathbf{c})^2\big), \nonumber\\
  \mathcal{L}_{\mathrm{NLL}} &= \tfrac{1}{2}\Bigl(\log 2\pi + 2\log\sigma
  + \tfrac{(y-\mu)^2}{\sigma^2}\Bigr). \label{ehm:eq:gen-mve}
\end{align}
\emph{where} $\mu(\cdot)$ and $\sigma(\cdot)\!>\!0$ are neural outputs, $\mathcal{L}_{\mathrm{NLL}}$ is the negative log-likelihood loss function with a given statistical assumption (i.e. Gaussian). A two-sided PI at level $1-\alpha$ is $[\mu - z_{1-\alpha/2}\sigma,\ \mu + z_{1-\alpha/2}\sigma]$ with $z_q$ the standard-normal quantile \cite{NixWeigend1994}.

\subsection{Exponential Degradation}
A common degradation law uses exponential growth (or decay):
\begin{align}
  s(t) &= A\,e^{B t}, \qquad t_\gamma \;=\; \frac{1}{B}\,\log\!\Bigl(\frac{\gamma}{A}\Bigr). \label{ehm:eq:gen-edm}
\end{align}
\emph{where} $s(t)$ is the indicator at operational age $t$, $A\!>\!0$ is amplitude, $B\!>\!0$ is rate, $\gamma\!>\!0$ is a threshold, and $t_\gamma$ is the first crossing time \cite{Gebraeel2005}.

\subsection{Weibull Survival}
Let $T$ denote the time-to-event. The Weibull survival, density, and hazard are, respectively,
\begin{align}
  S(t) &= \exp\!\Bigl[-(t/\lambda)^k\Bigr], \quad
  f(t) = \frac{k}{\lambda}\Bigl(\frac{t}{\lambda}\Bigr)^{k-1}\!\exp\!\Bigl[-(t/\lambda)^k\Bigr], \quad\\
  h(t) &= \frac{f(t)}{S(t)} = \frac{k}{\lambda}\Bigl(\frac{t}{\lambda}\Bigr)^{k-1}. \label{ehm:eq:gen-weibull}
\end{align}
With observed $(t_i,\delta_i)$, the right-censored log-likelihood is
\begin{align}
  \ell &= \sum_{i}\Bigl\{\delta_i \log f(t_i\mid k,\lambda) + (1-\delta_i)\log S(t_i\mid k,\lambda)\Bigr\}, \label{ehm:eq:gen-weibull-ll}
\end{align}
\emph{where} $k\!>\!0$ and $\lambda\!>\!0$ are shape and scale; $\delta_i\!=\!1$ for events, $0$ for censoring. Weibull modeling is standard in reliability \cite{Weibull1951,MeekerEscobar1998}.

\subsection{Evaluation Metrics for Uncertainty Quantification}
Given targets and prediction intervals (PIs) $\{(y_n,[\ell_n,u_n])\}_{n=1}^{N}$, evaluation metrics are defined as follows
\begin{align}
  \mathrm{PICP} &= \frac{1}{N}\sum_{n=1}^{N} \mathbf{1}\{\ell_n \le y_n \le u_n\}, \label{ehm:eq:picp}\\
  \mathrm{MPIW} &= \frac{1}{N}\sum_{n=1}^{N} (u_n-\ell_n), \label{ehm:eq:mpiw}\\
  \mathrm{NMPIW} &= \frac{1}{N}\sum_{n=1}^{N} \frac{(u_n-\ell_n)}{R}, \label{ehm:eq:nmpiw}
\end{align}
\noindent Here, $95\%$ PIs (i.e., using $z_{0.975}=1.96$ in \eqref{ehm:eq:mve-pi}) are used as a standard statistical convention.
\emph{where} $\mathrm{PICP}$ is the prediction interval coverage probability, $\mathrm{MPIW}$ is the mean PI width, $\mathrm{NMPIW}$ is the normalized width (with $R$ the observed target range), $\mathbf{1}\{\cdot\}$ is an indicator function, and $N$ is the sample size \cite{arik2015neural}.

The coverage width-based criterion (CWC) summarizes the sharpness--coverage tradeoff:
\begin{align}
  \mathrm{CWC}
  &= \mathrm{NMPIW}\,\Bigl[\,1+\alpha_c\,\exp\!\bigl(-\beta_c(\mathrm{PICP}-\gamma_{\mathrm{cov}})\bigr)\Bigr], \label{ehm:eq:cwc}
\end{align}
\noindent \emph{where} $\gamma_{\mathrm{cov}}\in(0,1)$ is a target coverage and $\alpha_c,\beta_c>0$ tune the penalty as coverage falls below the target \cite{Khosravi2011Review}.

\paragraph{Differentiable soft coverage surrogate}
To include coverage behavior directly in the training objective, the hard indicator is replaced in \eqref{ehm:eq:picp} with a smooth surrogate that is differentiable with respect to the interval bounds:
\begin{align}
  S(x) &:= \frac{1}{1+\exp(-x)},\\
  \widetilde{c}_n &= S\!\bigl(\tau (y_n-\ell_n)\bigr)\,S\!\bigl(\tau (u_n-y_n)\bigr), \label{ehm:eq:soft-coverage}\\
  \widetilde{\mathrm{PICP}} &= \frac{1}{N}\sum_{n=1}^{N}\widetilde{c}_n. \label{ehm:eq:soft-picp}
\end{align}
Here $\tau>0$ controls the smoothness (larger $\tau$ more closely approximates the hard indicator). Section~\ref{ehm:subsec:mtl} uses $\widetilde{\mathrm{PICP}}$ in a differentiable CWC regularizer that can be optimized via backpropagation.

\section{Predictive Performance Degradation with Scientific Machine Learning and Uncertainty Quantification}
\label{ehm:sec:Framework}

\begin{figure*}[!t]
    \centering
    \includegraphics[width=\textwidth]{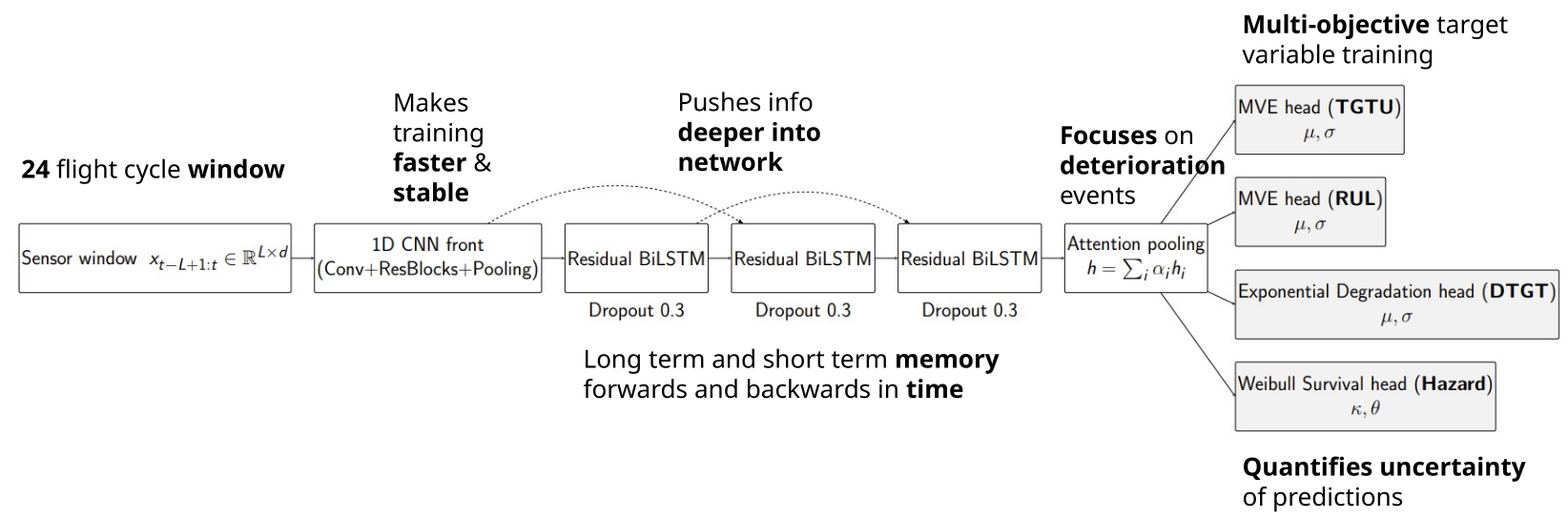}
    \caption{Novel multi-objective Engine Health Management neural network architecture.}
    \label{ehm:fig:Model-architecture}
\end{figure*}

\paragraph{Architectural overview}
Windows of multivariate sensor data are processed to produce uncertainty-aware predictions for turbine gas temperature (TGTU), Delta Turbine Gas Temperature (DTGT), remaining useful life (RUL), and degradation/survival quantities.
The proposed backbone consists of a 1-D convolutional front-end, a residual BiLSTM stack, and an attention pooling layer; multi-objective learning task heads of the network operate on the pooled context. \Cref{ehm:fig:Model-architecture} shows the neural network architecture.

\paragraph{Architecture design rationale}
The architecture is designed to balance expressiveness, robustness, and deployability on real fleet time series data. A convolutional front-end serves as a learnable feature extractor and down-sampler, helping suppress high-frequency noise and reduce sequence length before recurrent modeling. A residual bidirectional LSTM stack then captures both forward and backward temporal context within the window and improves optimization stability through skip connections \cite{Hochreiter1997,SchusterPaliwal1997,He2016ResNet}. Attention pooling summarizes the sequence into a context vector by focusing on the most informative time steps \cite{Bahdanau2015}, which is useful when relevant degradation signatures are intermittent and phase-dependent. Finally, a shared encoder with task-specific heads supports multi-task learning, improving sample efficiency relative to training separate models for each target and promoting representations that are consistent across related health indicators. Simpler encoders (e.g., a pure CNN or a single unidirectional RNN without attention) can be effective on benchmarks, but on heterogeneous operational data they may under-represent long-range dependencies or entangle regime effects; the modular design here provides a practical compromise between capacity and interpretability.

\begin{table*}[!t]
\centering
\small
\caption{Input and target variables.}
\label{ehm:tab:var_dictionary}
\begin{tabularx}{\textwidth}{>{\setlength\hsize{.17\hsize}} X >{\setlength\hsize{.15\hsize}} l >{\setlength\hsize{.68\hsize}} X}
\toprule
\textbf{Variable} & \textbf{Units} & \textbf{Description} \\
\midrule
\multicolumn{3}{l}{\textbf{Input variables}} \\
\addlinespace[2pt]
N1  & Percentage     & Low pressure spool speed (LP) \\
N2  & Percentage     & Intermediate pressure spool speed (IP) \\
N3  & Percentage     & High pressure spool speed (HP) \\
P25 & PSI            & Intermediate pressure compressor delivery pressure (IPC) \\
P3  & PSI            & High pressure compressor delivery pressure (HPC) \\
P50 & Pascal (Pa)    & Pressure \\
T2A & Kelvin (K)     & Temperature \\
T3  & Kelvin (K)     & Temperature \\
GWT & Kg             & Aircraft Gross Weight \\
MN  & N/A            & Marker number for the runway \\
\addlinespace
\multicolumn{3}{l}{\textbf{Select input variables (excluded from the RUL head to prevent leakage)}} \\
\addlinespace[2pt]
\makecell[l]{Cycles\\Since\\New}  & Cycles & The running total of accumulated flight cycles since the engine was new \\
\makecell[l]{Hours\\Since\\New}   & Hours  & The running total of accumulated flight hours since the engine was new \\
\addlinespace
\multicolumn{3}{l}{\textbf{Target variables}} \\
\addlinespace[2pt]
TGTU & Degrees Celsius & Turbine gas temperature untrimmed \\
DTGT & Degrees Celsius & TGTU -- nominal TGT \\
RUL  & Cycles          & Predicted cycles remaining before end of life (EOL) \\
Survival & Probability  & The predicted probability that the engine has not ``failed'' by the given time/cycle horizon \\
\bottomrule
\end{tabularx}
\end{table*}

\subsection{Convolutional Front-End}
Stacked Conv1D+ResBlocks map a lookback window to a shorter sequence using the operator in \eqref{ehm:eq:gen-cnn-resblock} and pooling length in \eqref{ehm:eq:gen-cnn-length}:
\begin{align}
  \mathbf{z}_i &= g_{\mathrm{conv}}\!\bigl(\mathbf{X}_{t-L+1:t}\bigr), \quad i=1,\dots,T'. \label{ehm:eq:conv}
\end{align}
\emph{where} $\mathbf{X}_{t-L+1:t}\!\in\!\mathbb{R}^{L\times d}$ is the window ending at $t$, $g_{\mathrm{conv}}(\cdot)$ denotes the concrete front-end (conv, BN, nonlinearity, pooling), $\mathbf{z}_i\!\in\!\mathbb{R}^{w}$ is the feature at reduced index $i$, and $T'$ is the post-pooling length.

To align the convolutional feature width $w$ with the BiLSTM hidden dimension $h$ for residual summation, optionally apply a pointwise linear projection
\begin{align}
  \tilde{\mathbf{z}}_i &= \mathbf{W}_{\mathrm{proj}}\,\mathbf{z}_i + \mathbf{b}_{\mathrm{proj}}, \qquad \tilde{\mathbf{z}}_i\in\mathbb{R}^{h},\; i=1,\dots,T'.
  \label{ehm:eq:proj}
\end{align}

\subsection{Residual BiLSTM Backbone}
Four BiLSTM layers with a residual connection read the sequence as in \eqref{ehm:eq:gen-bilstm-stack}:
\begin{align}
  \mathbf{h}^{(1)}_{1:T'} &= \mathrm{BiLSTM}_1(\tilde{\mathbf{z}}_{1:T'}), \notag\\
  \mathbf{h}^{(2)}_{1:T'} &= \mathrm{BiLSTM}_2(\mathbf{h}^{(1)}_{1:T'}) + \mathbf{h}^{(1)}_{1:T'},\\
  \mathbf{h}^{(3)}_{1:T'} &= \mathrm{BiLSTM}_3(\mathbf{h}^{(2)}_{1:T'}) + \mathbf{h}^{(2)}_{1:T'},\\
    \mathbf{h}^{(4)}_{1:T'} &= \mathrm{BiLSTM}_4(\mathbf{h}^{(3)}_{1:T'}) + \mathbf{h}^{(3)}_{1:T'}
  \label{ehm:eq:bilstm}
\end{align}
\emph{where} $\mathrm{BiLSTM}_\ell(\cdot)$ is a bidirectional LSTM with hidden size $h$, and $\mathbf{h}^{(\ell)}_{1:T'}$ denotes the hidden-state sequence output from BiLSTM layer $\ell\in\{1,\dots,4\}$.

\subsection{Attention Pooling}
The attention pooling layer follows \eqref{ehm:eq:gen-attn} to produce a context vector:
\begin{align}
  \alpha_i &= \frac{\exp(\mathbf{v}^{\top}\mathbf{h}^{(4)}_i)}{\sum_{j=1}^{T'} \exp(\mathbf{v}^{\top}\mathbf{h}^{(4)}_j)}, \quad \sum_{i=1}^{T'}\alpha_i=1, \label{ehm:eq:attn-w}\\
  \mathbf{c} &= \sum_{i=1}^{T'} \alpha_i\, \mathbf{h}^{(4)}_i. \label{ehm:eq:attn-c}
\end{align}
\emph{where} $\mathbf{v}\!\in\!\mathbb{R}^{h}$ is a learned attention vector, $\alpha_i$ are weights over $T'$ steps, and $\mathbf{c}\!\in\!\mathbb{R}^{h}$ is the pooled context \cite{Bahdanau2015}.

\subsection{Mean--Variance Estimation (MVE) Heads}
For any scalar regression target $y$ (TGTU, DTGT, or RUL), the Gaussian negative log-likelihood implied by \eqref{ehm:eq:gen-mve} is used to obtain head parameters by minimizing
\begin{align}
  \mathcal{L}_{\mathrm{NLL}}(\mu,\sigma;y)
  &= \frac{1}{2}\Bigl( \log(2\pi) + 2\log\sigma + \frac{(y-\mu)^2}{\sigma^2} \Bigr), \label{ehm:eq:mve}
\end{align}
\emph{where} $\mu=\mu(\mathbf{c})$ and $\sigma=\sigma(\mathbf{c})>0$ are head outputs conditioned on $\mathbf{c}$, and $y$ is the supervised target.
At confidence level $1-\alpha$, the analytic PI is
\begin{align}
  \ell &= \mu - z_{1-\alpha/2}\,\sigma, \quad
  u \;=\; \mu + z_{1-\alpha/2}\,\sigma. \label{ehm:eq:mve-pi}
\end{align}
\emph{where} $z_q$ is the standard normal quantile, and $\ell,u$ are the PI bounds.

%\subsection{Degradation Head (Performance, Rate, Time-to-Threshold)}
%A three-output MVE head produces degradation summaries:
%\begin{align}
%  (p,\sigma_p),\ (r,\sigma_r),\ (\tau,\sigma_\tau) &= g_{\mathrm{deg}}(\mathbf{c}). \label{ehm:eq:deg-head}
%\end{align}
%\emph{where} $p\in[0,1]$ is a normalized performance index, $r>0$ is a non-negative degradation rate, $\tau\ge 0$ is the predicted cycles to a threshold, and $\sigma_p,\sigma_r,\sigma_\tau$ are the associated standard deviations.

\subsection{Exponential Degradation Head}
The head follows the generic law in \eqref{ehm:eq:gen-edm}:
\begin{align}
  s(t) &= A\,e^{B t}, \qquad t_\gamma \;=\; \frac{1}{B}\,\log\!\Bigl(\frac{\gamma}{A}\Bigr). \label{ehm:eq:edm-ttl}
\end{align}
\emph{where} $s(t)$ is the indicator at age $t$, $A>0$ is amplitude, $B>0$ is rate, $\gamma>0$ is a threshold, and $t_\gamma$ is the crossing time \cite{Gebraeel2005}.

\subsection{Weibull Survival Head}
The network's survival head adopts \eqref{ehm:eq:gen-weibull}:
\begin{align}
  S(t) &= \exp\!\Bigl[-(t/\lambda)^k\Bigr],\quad
  h(t) \;=\; \frac{k}{\lambda}\Bigl(\frac{t}{\lambda}\Bigr)^{k-1}. \label{ehm:eq:weibull}
\end{align}
\emph{where} $k,\lambda>0$ are shape and scale, respectively.

\subsection{Multitask Objective with Learned Weights}
\label{ehm:subsec:mtl}
Let $\mathcal{T}$ denote the set of prediction heads (tasks) and let $L_k$ be the loss for task $k\in\mathcal{T}$.
For MVE regression heads (TGTU, DTGT, and RUL), the Gaussian negative log-likelihood (NLL) is minimized in \eqref{ehm:eq:mve} and augmented with a differentiable coverage--width regularizer:
\begin{align}
  L_k &= L_{\mathrm{NLL},k} + \lambda_{\mathrm{cwc}}\,\mathrm{CWC}^{\mathrm{soft}}_k,
  \qquad k\in\mathcal{T}_{\mathrm{MVE}}, \label{ehm:eq:taskloss}
\end{align}
\noindent \emph{where} $\lambda_{\mathrm{cwc}}\ge 0$ is a tunable weight, $\mathcal{T}_{\mathrm{MVE}}=\{\mathrm{TGTU},\mathrm{DTGT},\mathrm{RUL}\}$, and $\mathrm{CWC}^{\mathrm{soft}}_k$ is a differentiable analogue of the CWC metric in \eqref{ehm:eq:cwc}.
Using the PI bounds from \eqref{ehm:eq:mve-pi}, the mini-batch soft coverage is computed as $\widetilde{\mathrm{PICP}}_k$ using \eqref{ehm:eq:soft-picp} (a smooth approximation to \eqref{ehm:eq:picp}) and the normalized mean PI width $\mathrm{NMPIW}_k$ using \eqref{ehm:eq:nmpiw}. The regularizer is then
\begin{align}
  \mathrm{CWC}^{\mathrm{soft}}_k
  &= \mathrm{NMPIW}_k\,\Bigl[\,1+\alpha_c\,\exp\!\bigl(-\beta_c(\widetilde{\mathrm{PICP}}_k-\gamma_{\mathrm{cov}})\bigr)\Bigr]. \label{ehm:eq:cwc-soft}
\end{align}
Because $\widetilde{\mathrm{PICP}}_k$ is differentiable with respect to the PI bounds $(\ell_{n,k},u_{n,k})$, the CWC term can be backpropagated and optimized jointly with the NLL.

Finally, task losses are combined using learned uncertainty weights \cite{KendallGal2018}:
\begin{align}
  \mathcal{L}_{\mathrm{MTL}} &= \sum_{k\in\mathcal{T}}\Bigl( e^{-s_k} L_k + s_k \Bigr), \label{ehm:eq:mtl}
\end{align}
\noindent \emph{where} $s_k$ are learned log-variance parameters (larger $s_k$ down-weights task $k$).

\section{Experiment Setup}
\label{ehm:sec:Setup}

\paragraph{Optimizer and learning rate scheduler}
Training is conducted with AdamW and cosine scheduling \cite{loshchilov2016sgdr}; gradient clipping stabilizes the BiLSTM stack; hyperparameters (number of layers, $h$, dropout, attention) are selected by a manual, iterative investigation of network configurations not detailed in this paper.

\paragraph{Reproducibility} Data splits are logged, scalers are serialized, and best performing models are saved as pickle files.

\paragraph{Dataset splitting} One hundred twenty-two engines comprise the overall dataset. For each engine, data are available in three different flight phases: takeoff, climb, and cruise. Additionally, for each flight phase, there are various flight segments ranging from one to six. In other words, the overall time series of flight records is partitioned by scheduled maintenance overhauls. This partitioning is what creates the segments.

\begin{figure}[H]
    \centering
    \includegraphics[width=\linewidth]{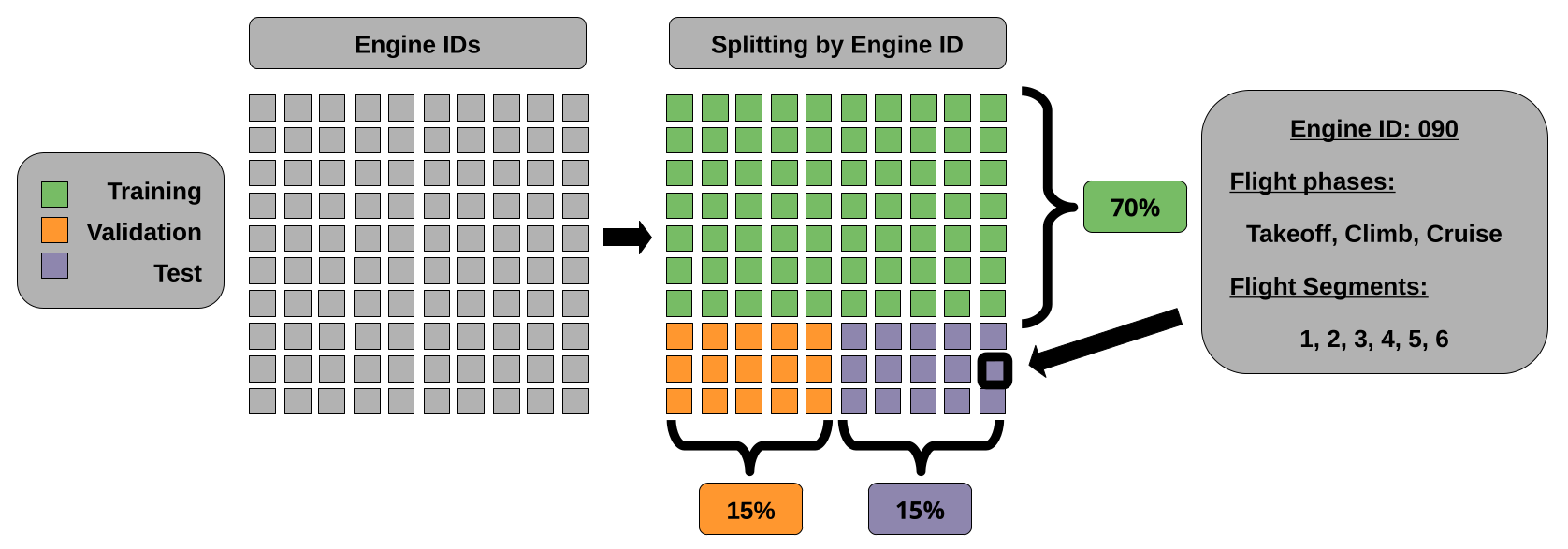}
    \caption{Engine Health Management data are split by unique engine serial number using a ratio of 70\%, 15\%, and 15\% for training, validation, and test datasets, respectively.}
    \label{ehm:fig:Data-splitting}
\end{figure}

Training, validation, and test datasets are split by unique engine ID. Splitting in this manner avoids data leakage, since all flight phases and flight segments for one given engine stay within a given dataset split. \Cref{ehm:fig:Data-splitting} illustrates how the data are split using a 70/15/15 ratio for training, validation, and test, respectively. Specifically, this translates to 85 engine IDs used for training, 18 engine IDs used for validation, and 19 engine IDs used for testing (remainders are allocated to the test dataset). For simplicity, rows with missing data are removed. Data imputation methods are explored but are not detailed in this paper. The input and target variables are detailed in \cref{ehm:tab:var_dictionary}. Select input variables are not passed into the RUL head of the network to prevent data leakage.

Robust data scaling is used as a preprocessing step. Scaling data helps the network converge on solutions faster and improves predictive performance. Robust scaling is used because it is based on percentiles and consequently is not impacted by a small number of large outliers, unlike scaling based on minimum and maximum values \cite{scikit-learn, sklearn_api}.

After robust scaling, each sensor channel is detrended using a (linear) Kalman filter with a constant-velocity state model to separate a smooth trend from higher-frequency residuals \cite{kalman1960, lacey1998tutorial}. Let the latent trend state be $\mathbf{m}_t=[p_t\; v_t]^\top$ (position and velocity) and the measurement be $y_t$, thus giving
\begin{align}
  \mathbf{m}_t &= \mathbf{F}\,\mathbf{m}_{t-1} + \mathbf{w}_t, \qquad y_t = \mathbf{H}\,\mathbf{m}_t + \nu_t, \label{ehm:eq:kf-detrend}
\end{align}
with $\mathbf{F}=\begin{bmatrix}1 & \Delta t\\ 0 & 1\end{bmatrix}$ and $\mathbf{H}=[1~~0]$. The detrended value is defined as $\tilde y_t = y_t - \hat p_t$, where $\hat p_t$ is the filtered estimate of $p_t$.

After detrending, the data are reformatted into time-series sequences, with a sliding window consisting of twenty-four observations per window, and an overlapping stride of one observation \cite{perea2013slidingwindowspersistenceapplication}. One data observation is equivalent to one flight cycle. A window size of twenty-four was chosen to balance the amount of short-term and long-term trends the network sees during training. A stride of one was chosen to maximize the number of sequences created for network training purposes. Next, DTGT thresholds are investigated for the purposes of creating hazard events as a target variable. This created target variable is used to train the Weibull hazard head of the network, whose hazard function is given in \eqref{ehm:eq:weibull}.

\paragraph{Data-driven DTGT thresholding for Weibull survival}
First, it is important to establish a \emph{healthy} baseline for each engine--phase--segment by taking a prefix consisting of the first $100$ samples or the first $20\%$ of the segment---whichever is smaller. Let $T$ be the segment length and define the prefix $\mathcal{H}=\{x_t\}_{t=1}^{L}$ with $L=\min\!\left(100,\lceil 0.2\,T\rceil\right)$. From $\mathcal{H}$, robust location and scale are computed: the center is the median $\tilde{\mu}=\operatorname{median}(\mathcal{H})$, and the spread is the median absolute deviation (MAD) rescaled to a standard-deviation equivalent,
\begin{equation}
\label{ehm:eq:robust-mad-scale}
\hat{\sigma}=1.4826 \cdot \operatorname{median}\{|x-\tilde{\mu}|: x\in\mathcal{H}\}.
\end{equation}
a choice that is resistant to outliers yet statistically consistent under normality \cite{shewhart1923_economic_control,shewhart_deming1939_viewpoint_qc,montgomery2005_ISQC5e,nist2016_mad}.

Next, a \emph{robust Shewhart-style control limit} for detection is formed. The DTGT threshold is
\begin{equation}
\label{ehm:eq:dtgt-threshold}
\mathrm{thr} \;=\; \max\!\left(\mathrm{floor},\, \tilde{\mu} + k\,\hat{\sigma}\right),
\end{equation}
where $k>0$ is a tunable sigma-multiplier (similarly to the ``$k\sigma$'' idea from Statistical Process Control (SPC)) and \textit{floor} is a fixed lower bound that prevents the threshold from collapsing toward zero on extremely quiet baselines. This is the robust analogue of a ``mean $+\,k\sigma$'' Shewhart limit computed from the healthy baseline of the signal after scheduled service (the nominal engine) while adding a quiet-floor safeguard \cite{shewhart1923_economic_control,montgomery2005_ISQC5e,nist_esh_631_control_charts}.

To convert threshold exceedances into events suitable for survival modeling, a \emph{runs-rule} is applied. A ``hazard'' is declared at the first time index $t_e$ at which there are $m$ consecutive samples above the threshold,
\begin{equation}
\label{ehm:eq:runs-rule}
\prod_{j=0}^{m-1}\left(x_{t_e-j}>\mathrm{thr}\right)=1,
\end{equation}
and no hazard is recorded if such a run never occurs. This mirrors the Western Electric runs tests that are used on control charts to raise specificity and suppress spurious single-point alarms near the limit \cite{western1956_sqc_handbook,qimacros_nelson_rules}.

Finally, a \emph{preprocessing parameter sweep} over the three knobs $(k,\mathrm{floor},m)$ is performed. For each combination, three diagnostic metrics are computed. First, the \emph{Health Exceedance Rate (HER)} is the fraction of prefix samples above the threshold,
\begin{equation}
\label{ehm:eq:health-exceedance-rate}
\mathrm{HER}=\frac{1}{L}\sum_{t=1}^L \mathds{1}\!\left(x_t>\mathrm{thr}\right).
\end{equation}
Second, \texttt{event\_\allowbreak windows\_\allowbreak pct} is the proportion of windows belonging to an event run. Third, \texttt{segments\_\allowbreak event\_\allowbreak pct} is the proportion of segments containing at least one event (first crossing). \Cref{ehm:table:threshold-variables} provides more detail on these variables. A desirable result balances the false-alarm control against the number of usable positives for learning. These statistics allow practitioners to select thresholds consistent with in-house preferences for conservatism versus sensitivity while ensuring the survival model receives enough positive signal to learn \cite{shewhart1923_economic_control,western1956_sqc_handbook,montgomery2005_ISQC5e,nist_esh_631_control_charts,qimacros_nelson_rules}.

\paragraph{Parameter-sweep analysis}
Using the sweep table shown in \cref{ehm:table:threshold-sweep}, the \emph{floor} parameter dominates: for this dataset, any floor value $\ge 5$ eliminates detections entirely, meaning HER $= 0\%$, \texttt{event\_\allowbreak windows\_\allowbreak pct} $= 0$, and \texttt{segments\_\allowbreak event\_\allowbreak pct} $= 0$. That ``quiet floor'' behavior aligns with SPC intuition: fixed lower limits can suppress signals from small but genuine shifts near the baseline, as emphasized in Shewhart-style control and run-rules guidance. With \emph{floor} $= 0$, the sigma-multiplier $k$ trades off false alerts in the healthy prefix against the prevalence of events. In these data, $k=4.0$ yields a HER $\approx 0.10\%$, \texttt{failures\_\allowbreak pct\_\allowbreak legacy} $\approx 8.39\%$, \texttt{event\_\allowbreak windows\_\allowbreak pct} $\approx 0.0139$, and \texttt{segments\_\allowbreak event\_\allowbreak pct} $\approx 33.33\%$, making for a balanced operating point with low false alarms yet enough positives for the Weibull model to learn. A looser $k=3.5$ increases HER and event rates (more positives), whereas a stricter $k=4.5$ reduces them (fewer positives). This knob, together with the consecutive-run length $m$ (a Western Electric--style rule), provides an interpretable, tunable way to align detection specificity with in-house policy and quality-control doctrine rooted in Shewhart control chart, Statistical Process Control, and run-rules practice.

\begin{algorithm}[!t]
\small
\DontPrintSemicolon
\SetAlgoLined
\SetKwInOut{Input}{Input}
\SetKwInOut{Params}{Parameters}
\SetKwInOut{Output}{Output}

\caption{Creation of Target Variables: Robust DTGT Threshold \& Hazard Event}

\Input{Time series $x_{1:T}$ of DTGT.}
\Params{%
$k$ (sigma multiplier); \;
$m$ (consecutive run length, e.g., $m{=}3$); \;
$\textit{floor}$ (absolute threshold minimum); \;
$\textit{first\_n}$ (default $100$); \;
$\varphi$ (healthy prefix fraction, default $0.2$).}
\Output{Threshold $thr$; hazard flag $\texttt{hazard}\in\{\texttt{true},\texttt{false}\}$; first hazard time $t_e$ (or \texttt{None}).}

\BlankLine
\textbf{Healthy prefix length:} $L \leftarrow \min(\textit{first\_n}, \lceil \varphi T\rceil)$; \;
$\mathcal{H} \leftarrow \{x_t\}_{t=1}^{L}$ \tcp*[r]{prefix assumed ``healthy''}

\textbf{Robust center:} $\tilde{\mu} \leftarrow \mathrm{median}(\mathcal{H})$;\\
\textbf{MAD:} $\mathrm{MAD} \leftarrow \mathrm{median}_{x\in\mathcal{H}}\{|x-\tilde{\mu}|\}$;\\
\textbf{Scale to $\sigma$:} $\hat{\sigma} \leftarrow 1.4826 \times \mathrm{MAD}$ \tcp*[r]{normal-consistency}

\textbf{Threshold:} $thr \leftarrow \max\!\big(\textit{floor},\; \tilde{\mu} + k\,\hat{\sigma}\big)$ \;

\BlankLine
$\texttt{hazard}\leftarrow \texttt{false}$; \ $t_e \leftarrow \texttt{None}$ \;

\For{$t \leftarrow m,\ldots, T$}{
  \If{$\prod_{j=0}^{m-1} \mathbf{1}\!\left(x_{t-j} > thr\right) = 1$}{
     $\texttt{hazard}\leftarrow \texttt{true}$; \ $t_e\leftarrow t$; \ \textbf{break}\;
  }
}
\Return $thr$, $\texttt{hazard}$, $t_e$ \;
\end{algorithm}

\begin{table*}[!t]
\centering
\caption{Target variables for DTGT thresholding and hazard events.}
\label{ehm:table:threshold-variables}
\begin{tabularx}{\textwidth}{l l X}
\toprule
\textbf{Variable} & \textbf{Unit} & \textbf{Description} \\
\midrule
HER & Percentage &
Health Exceedance Rate. \% of DTGT values \emph{inside the healthy prefix window}
that exceed the DTGT threshold. \\

\texttt{failures\_pct\_legacy} & Percentage &
\% of windows flagged by legacy code (Boolean logic that marks hazard windows
after the first hazard occurrence; useful for masking). \\

\texttt{event\_windows\_pct} & Percentage &
\% of all windows that belong to an event run (i.e., any of the $m$
successive points that satisfy the rule). \\

\texttt{segments\_event\_pct} & Percentage &
\% of segments that contain at least one event (first--crossing) under the
current $(k,\ \textit{floor},\ m)$ sweep. \\
\bottomrule
\end{tabularx}
\end{table*}

\begin{table*}[!t]
\centering
\caption{DTGT threshold preprocessing parameter sweep results.}
\label{ehm:table:threshold-sweep}
\begin{adjustbox}{max width=\textwidth}
\scriptsize
\begin{tabular}{@{}cccccccccc@{}}
\toprule
$k$ & floor & HER & \shortstack{failures\\pct\\legacy} & \shortstack{event\\windows\\pct} &
\shortstack{segments\\event\\pct} & \shortstack{thr\\min} & \shortstack{thr\\median} & \shortstack{thr\\mean} & \shortstack{thr\\max}\\
\midrule
3.5 & 0  & 0.13 & 10.82 & 0.0139 & 36.30 & 0 & 0.76 & 0.84 & 2.91\\
3.5 & 5  & 0    & 0     & 0      & 0     & 5 & 5.00 & 5.00 & 5.00\\
3.5 & 8  & 0    & 0     & 0      & 0     & 8 & 8.00 & 8.00 & 8.00\\
3.5 & 10 & 0    & 0     & 0      & 0     & 10 & 10.00 & 10.00 & 10.00\\
3.5 & 12 & 0    & 0     & 0      & 0     & 12 & 12.00 & 12.00 & 12.00\\
\midrule
\textbf{4.0} & \textbf{0}  & \textbf{0.10} & \textbf{8.39} & \textbf{0.0139} &
\textbf{33.33} & \textbf{0} & \textbf{0.87} & \textbf{0.92} & \textbf{3.05}\\
4.0 & 5  & 0    & 0     & 0      & 0     & 5 & 5.00 & 5.00 & 5.00\\
4.0 & 8  & 0    & 0     & 0      & 0     & 8 & 8.00 & 8.00 & 8.00\\
4.0 & 10 & 0    & 0     & 0      & 0     & 10 & 10.00 & 10.00 & 10.00\\
4.0 & 12 & 0    & 0     & 0      & 0     & 12 & 12.00 & 12.00 & 12.00\\
\midrule
4.5 & 0  & 0.07 & 6.33 & 0.0141 & 29.63 & 0 & 0.99 & 1.00 & 3.21\\
4.5 & 5  & 0    & 0     & 0      & 0     & 5 & 5.00 & 5.00 & 5.00\\
4.5 & 8  & 0    & 0     & 0      & 0     & 8 & 8.00 & 8.00 & 8.00\\
4.5 & 10 & 0    & 0     & 0      & 0     & 10 & 10.00 & 10.00 & 10.00\\
4.5 & 12 & 0    & 0     & 0      & 0     & 12 & 12.00 & 12.00 & 12.00\\
\bottomrule
\end{tabular}
\end{adjustbox}
\end{table*}

\begin{figure*}[!t]
    \centering
    \includegraphics[width=\textwidth]{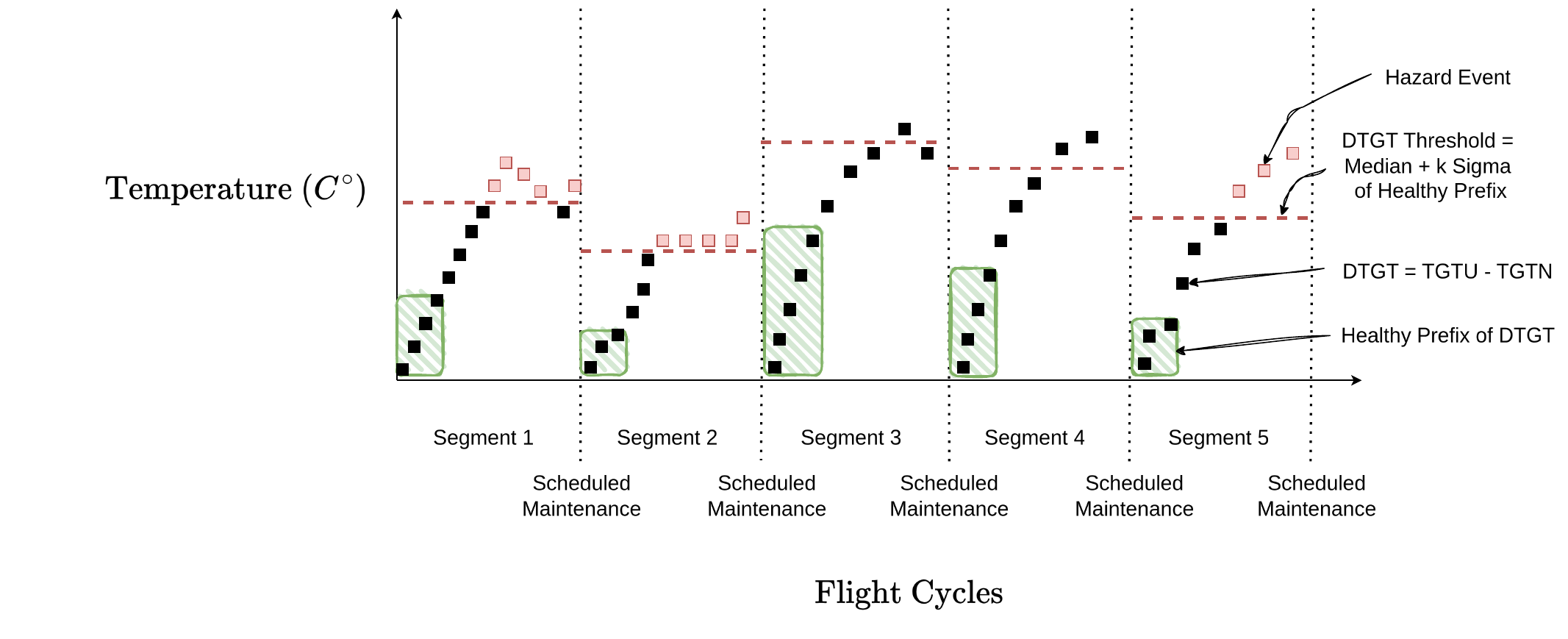}
    \caption{Deterioration thresholds and hazard events are created using data-driven statistical process control. Parameterized target variables enable in-house adjustment to proprietary threshold information.}
    \label{ehm:fig:Threshold-creation}
\end{figure*}

\begin{figure}[H]
    \centering
    \includegraphics[width=\linewidth]{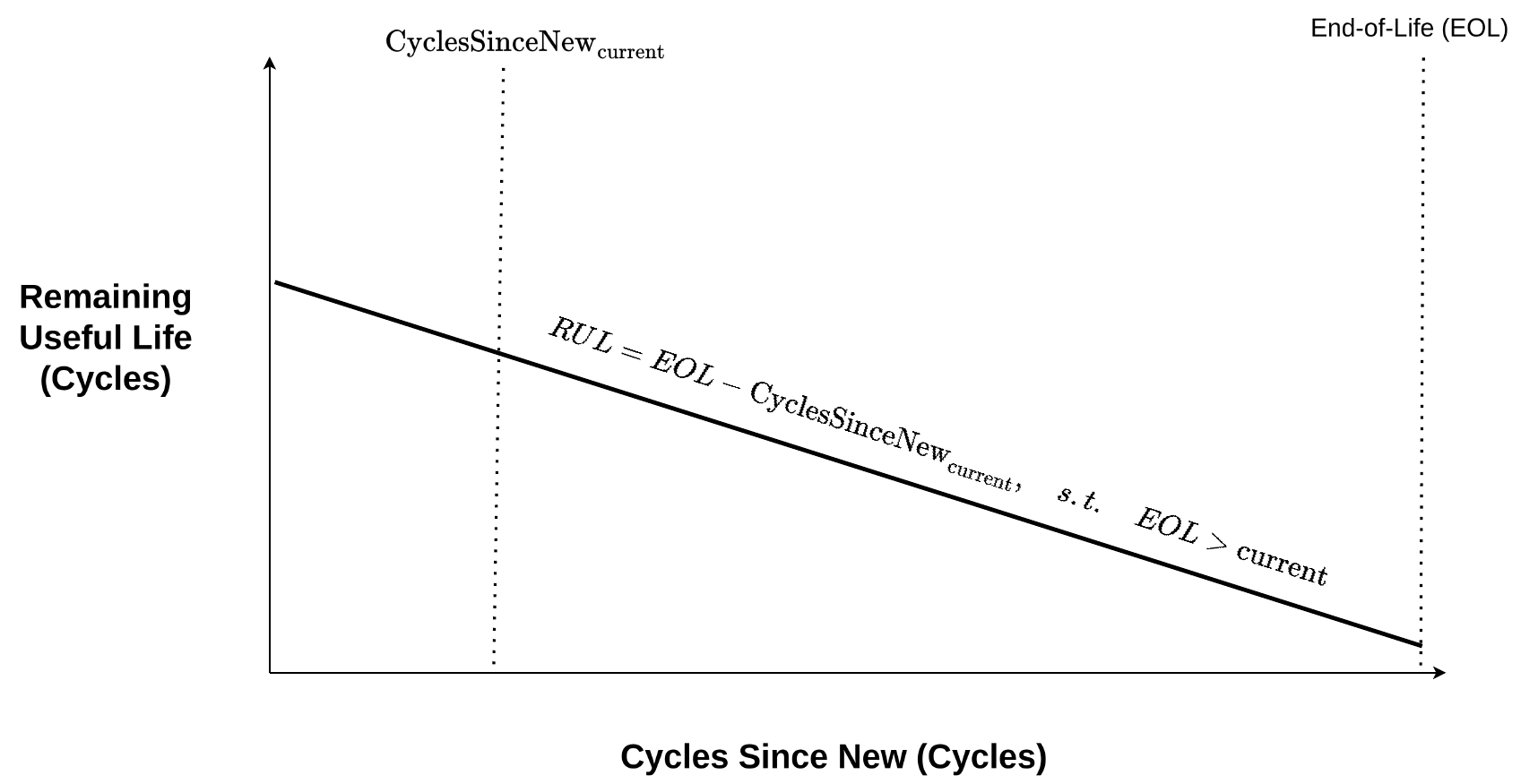}
    \caption{Creation of Remaining Useful Life (RUL) target variables. Parameterized target RUL variables are tunable per in-house specifications.}
    \label{ehm:fig:RUL-creation}
\end{figure}

\paragraph{RUL target construction}
For each engine/phase/segment trajectory, the end-of-segment index $t_{\mathrm{EOL}}$ is defined as the final observed cycle before the next scheduled maintenance/overhaul. A simple and commonly used target is the remaining number of cycles until that event,
\begin{align}
  \mathrm{RUL}_t &= t_{\mathrm{EOL}} - t, \qquad t=1,\dots,t_{\mathrm{EOL}},
  \label{ehm:eq:rul-def}
\end{align}
optionally clipped or re-parameterized to match in-house policy constraints (\Cref{ehm:fig:RUL-creation}).

\begin{figure}[H]
    \centering
    \includegraphics[width=0.94\columnwidth]{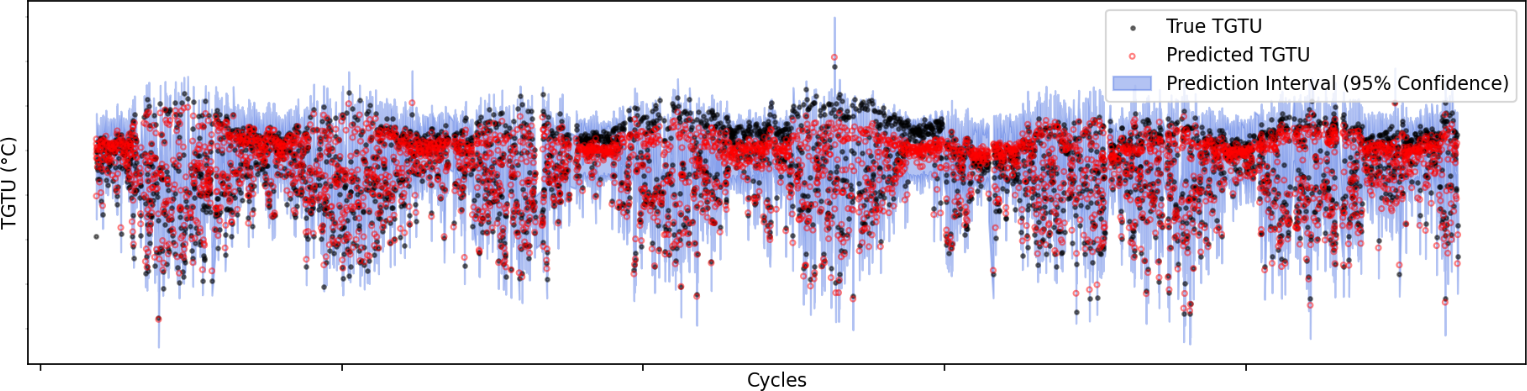}
    \caption{Predicted TGTU versus Engine Health Management TGTU (ground truth) with prediction intervals for engine B's climb phase during segment 4.}
    \label{ehm:fig:tgt-timeseries}
\end{figure}

\begin{figure}[H]
    \centering
    \includegraphics[width=0.94\columnwidth]{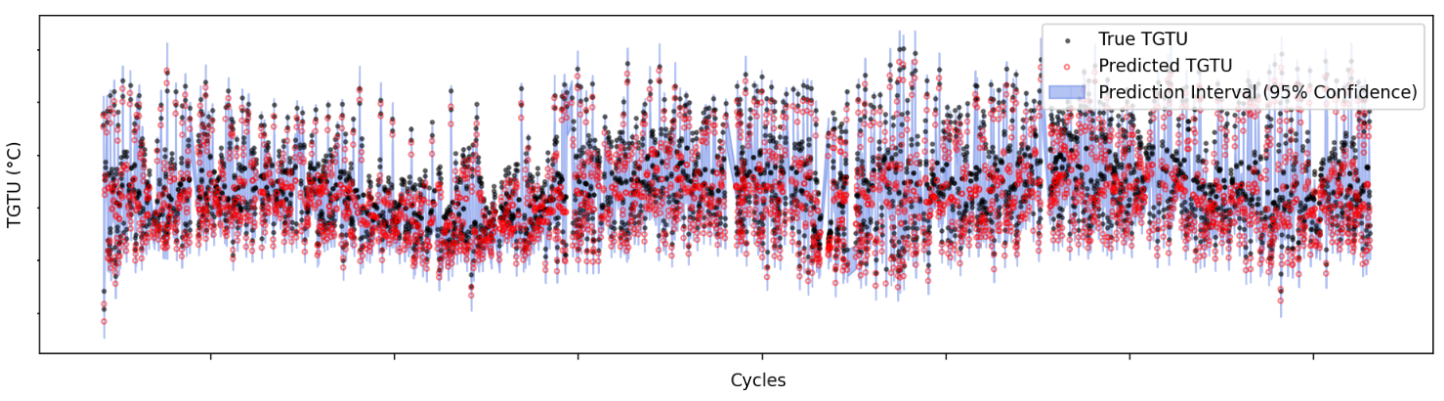}
    \caption{Predicted TGTU versus Engine Health Management TGTU (ground truth) with prediction intervals for engine S's takeoff phase during segment 3.}
    \label{ehm:fig:tgtu-timeseries-S-takeoff}
\end{figure}

\begin{figure}[H]
    \centering
    \includegraphics[width=0.94\columnwidth]{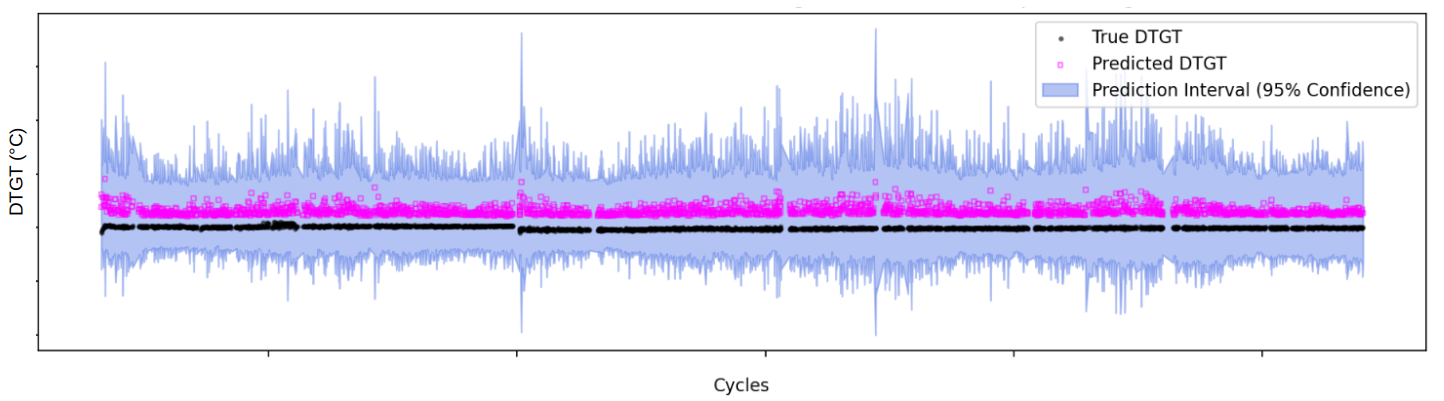}
    \caption{Predicted DTGT versus Engine Health Management DTGT (ground truth) with prediction intervals for engine B's cruise phase during segment 4.}
    \label{ehm:fig:dtgt-timeseries-B-cruise}
\end{figure}

\begin{figure}[H]
    \centering
    \includegraphics[width=0.94\columnwidth]{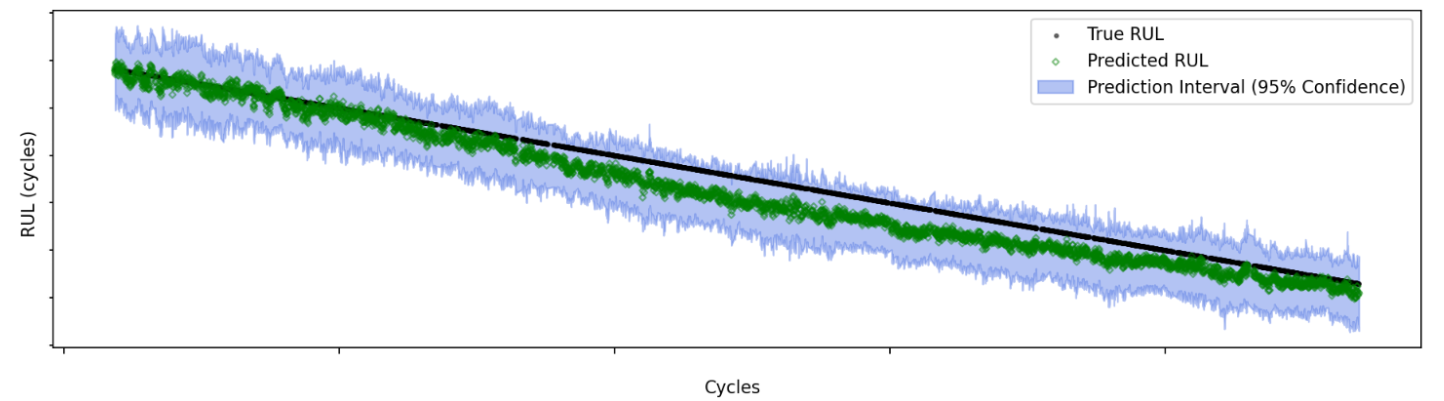}
    \caption{Predicted RUL versus Engine Health Management RUL (ground truth) with prediction intervals for engine B's cruise phase during segment 4.}
    \label{ehm:fig:rul-timeseries-B}
\end{figure}

\begin{figure}[H]
    \centering
    \includegraphics[width=0.94\columnwidth]{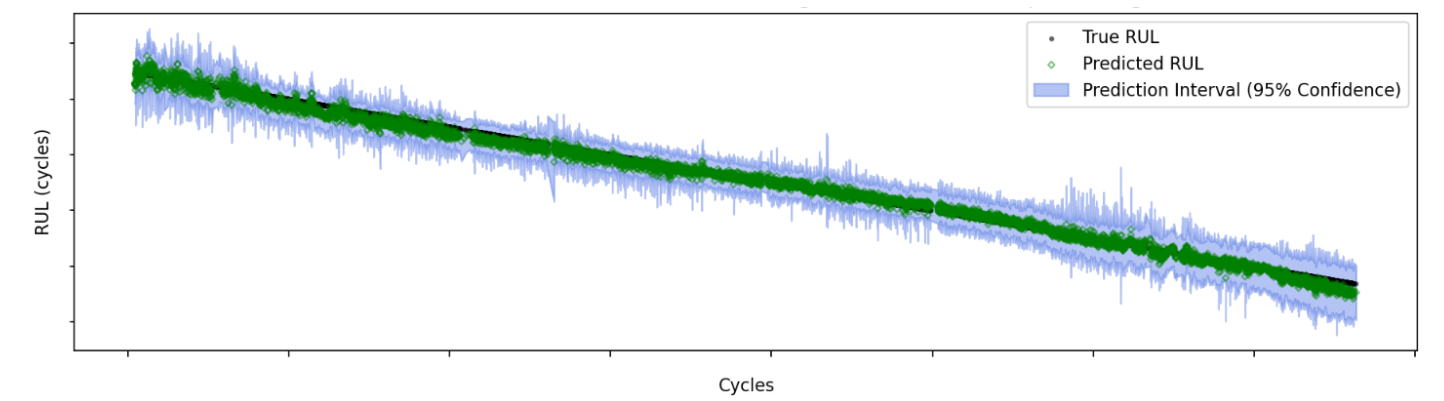}
    \caption{Predicted RUL versus Engine Health Management RUL (ground truth) with prediction intervals for engine O's takeoff phase during segment 2.}
    \label{ehm:fig:rul-timeseries-O-takeoff}
\end{figure}

\begin{figure}[H]
    \centering
    \includegraphics[width=0.94\columnwidth]{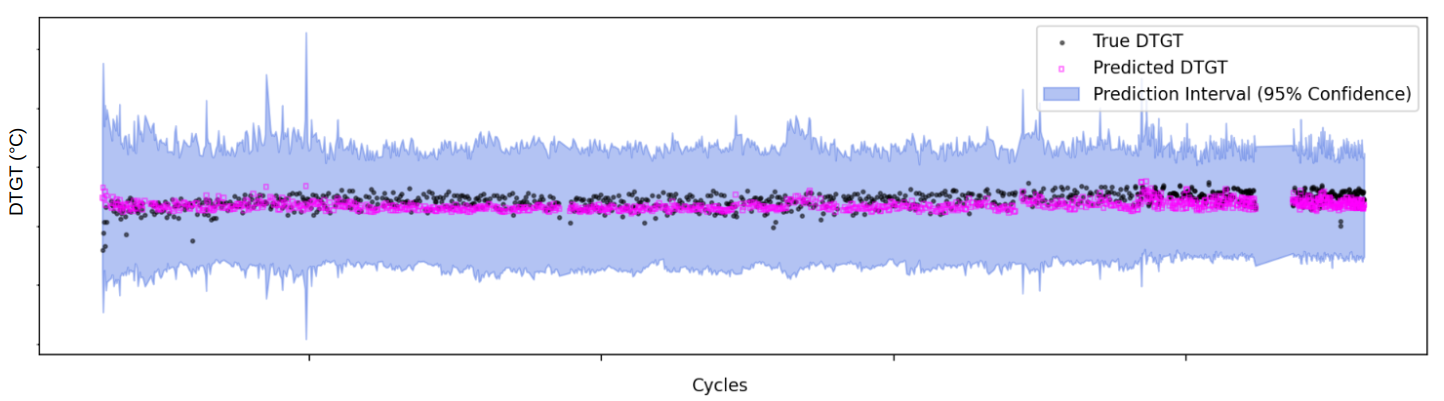}
    \caption{Predicted DTGT versus Engine Health Management DTGT (ground truth) with prediction intervals for engine S's takeoff phase during segment 2.}
    \label{ehm:fig:dtgt-timeseries-S-takeoff}
\end{figure}

\begin{figure}[H]
    \centering
    \includegraphics[width=0.94\columnwidth]{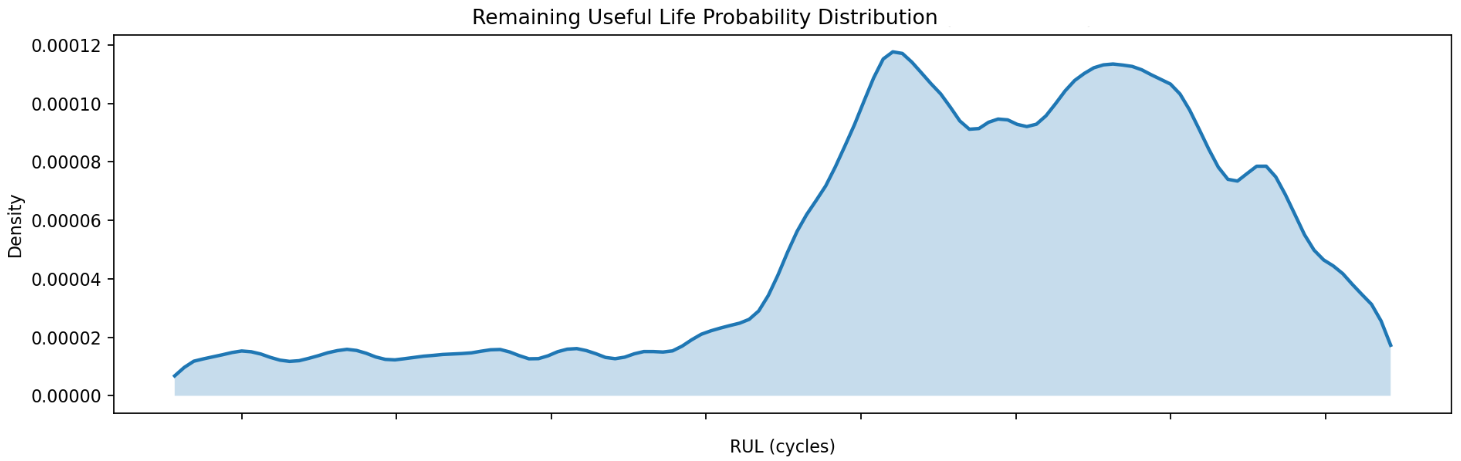}
    \caption{Remaining Useful Life Probability Distribution.}
    \label{ehm:fig:rul-pdf}
\end{figure}

%%%%%%%%%%%%%%%%%%%%%%%%%%%%%%%%%%%%%%%%%%%%%%%%%%%%%%%%%%%%%%%%%%%%%%
\section{Results}
\label{ehm:sec:Results}

Across the board, the presented framework produces excellent CWC evaluation metrics, with values often falling below one. However, the median CWC (MED-CWC) values reported in \cref{ehm:tbl:by_segment,ehm:tbl:by_esn_phase,ehm:tbl:tgtu_by_engine_phase,ehm:tbl:rul_by_engine_phase,ehm:tbl:tgtu_by_engine_phase_segment} occasionally take on very large magnitudes (spanning multiple orders of magnitude). This behavior follows directly from the CWC definition in \cref{ehm:eq:cwc}: CWC scales an interval-width term by an exponential penalty when empirical coverage (PICP) falls below the target coverage $\gamma$ (set by the chosen PI level). When results are stratified by flight phase, segment, and/or engine ID, the sample size per subset can be much smaller than the full test set, so PICP can deviate from $\gamma$ due to only a modest number of misses; the exponential term in \cref{ehm:eq:cwc} can then amplify that deviation into orders-of-magnitude increases in CWC.

This effect is visible already in the segment-level summary in \cref{ehm:tbl:by_segment}, where TGTU in segments~1 and~6 combines relatively small normalized widths (NMPIW $\approx 0.14$--$0.15$) with reduced coverage (PICP $0.76$ and $0.71$), producing MED-CWC values of \num{4.4e5} and \num{2.3e4}. The engine-level stratifications in \cref{ehm:tbl:by_esn_phase,ehm:tbl:tgtu_by_engine_phase} show that such outliers are concentrated in specific engines and phases (often takeoff and climb), including multiple engine--phase pairs with PICP far below nominal coverage. The most granular breakdown in \cref{ehm:tbl:tgtu_by_engine_phase_segment} further localizes these cases to specific engine/phase/segment combinations, confirming that the largest MED-CWC penalties align with the lowest-PICP subsets.

Importantly, once the exponential penalty dominates, the absolute magnitude of MED-CWC is best interpreted as a \emph{calibration-failure indicator} rather than a physically meaningful scale: differences between, say, \num{1e12} and \num{1e15} mainly reflect how far a subset's coverage falls below $\gamma$, not subtle changes in interval sharpness. Moreover, because \emph{median} CWC is reported, a large MED-CWC typically indicates that under-coverage is systematic for that subset (not driven by a single anomalous prediction). In practice, these outliers suggest localized miscalibration and/or a distribution shift in those operating conditions (e.g., strong transients, uncommon regimes, or sensor/segmentation artifacts). For deployment, these results motivate monitoring MED-CWC alongside the component metrics (PICP and MPIW/NMPIW) and applying targeted recalibration or investigation for engine/phase/segment combinations with persistently low coverage.

\Cref{ehm:tbl:overall} summarizes aggregate predictive accuracy and interval quality across the three tasks. The model attains mean absolute errors of $6.00\,^{\circ}\mathrm{C}$ for DTGT and $6.84\,^{\circ}\mathrm{C}$ for TGTU, and $531$ cycles for RUL. Empirical coverage (PICP) ranges from $0.86$ to $0.98$, while MPIW/NMPIW and MED-CWC quantify how wide the PIs must be to achieve that coverage.

\begin{table}[!t]
\centering
\caption{Overall Model Performance Metrics}
\label{ehm:tbl:overall}
\small

\begin{tblr}{
    width   = \linewidth,
    colspec = {X[l] *{5}{X[r]}},
    colsep  = 3pt
}
\toprule
Task & MAE & PICP & MPIW & NMPIW & CWC \\
\midrule
DTGT (\textdegree C) & 6.00 & 0.94 & 30.08 & 1.90 & 1.84 \\
RUL (cycles)         & 531  & 0.98 & 2569  & 1.18 & 1.15 \\
TGTU (\textdegree C) & 6.84 & 0.86 & 28.00 & 0.16 & 0.24 \\
\bottomrule
\end{tblr}
\end{table}

\Cref{ehm:tbl:by_phase} breaks these metrics down by flight phase, revealing phase-dependent difficulty. Takeoff exhibits the largest temperature errors (TGTU MAE $8.39\,^{\circ}\mathrm{C}$, DTGT MAE $7.23\,^{\circ}\mathrm{C}$) and the lowest TGTU coverage (PICP $0.79$), consistent with strong transients; cruise yields the lowest TGTU MAE ($5.00\,^{\circ}\mathrm{C}$) but the largest RUL MAE ($681$ cycles). These trends motivate phase-aware calibration and/or loss weighting.

\begin{table*}[!t]
\centering
\caption{Performance Metrics by Flight Phase}
\label{ehm:tbl:by_phase}
\small
\begin{tblr}{
  width   = \textwidth,
  colspec = {l l *{5}{X[r]}},
  colsep  = 3pt
}
\toprule
Phase & Task & MAE & PICP & MPIW & NMPIW & CWC \\
\midrule
\SetCell[r=3]{l} Takeoff
  & TGTU (\textdegree C) & 8.39 & 0.79 & 28.66 & 0.11 & 0.49 \\
  & RUL (cycles)         & 523  & 0.97 & 2305  & 0.91 & 1.18 \\
  & DTGT (\textdegree C) & 7.23 & 0.90 & 32.00 & 0.80 & 1.74 \\
\SetCell[r=3]{l} Climb
  & TGTU (\textdegree C) & 6.43 & 0.89 & 26.11 & 0.18 & 0.28 \\
  & RUL (cycles)         & 406  & 1.00 & 2444  & 1.79 & 0.88 \\
  & DTGT (\textdegree C) & 5.54 & 0.95 & 29.45 & 0.66 & 1.28 \\
\SetCell[r=3]{l} Cruise
  & TGTU (\textdegree C) & 5.00 & 0.94 & 29.11 & 0.19 & 0.21 \\
  & RUL (cycles)         & 681  & 0.97 & 3099  & 0.90 & 2.20 \\
  & DTGT (\textdegree C) & 4.71 & 1.00 & 27.93 & 4.89 & 7.01 \\
\bottomrule
\end{tblr}
\end{table*}

\Cref{ehm:tbl:by_segment} further decomposes performance by flight segment index, showing that both accuracy and calibration vary across segments. For example, TGTU MAE ranges from $4.95\,^{\circ}\mathrm{C}$ (segment~5) to $10.14\,^{\circ}\mathrm{C}$ (segment~6), and segments~1 and~6 exhibit low TGTU PICP ($0.76$ and $0.71$) with very large MED-CWC penalties, indicating that interval calibration for TGTU is particularly challenging in those segments.

\begin{table*}[!t]
\centering
\caption{Performance Metrics by Flight Segment}
\label{ehm:tbl:by_segment}
\small
% \small % optional
\begin{tblr}{
  width   = \textwidth,
  colspec = {l l *{5}{X[r]}},
  colsep  = 3pt
}
\toprule
Segment & Task & MAE & PICP & MPIW & NMPIW & CWC \\
\midrule
\SetCell[r=3]{l} 1
  & TGTU (\textdegree C) & 7.59 & 0.76 & 25.06 & 0.15 & \num{4.42e5} \\
  & RUL (cycles)         & 446  & 0.99 & 3233  & 2.06 & 1.52 \\
  & DTGT (\textdegree C) & 6.13 & 0.91 & 27.46 & 1.54 & 2.05 \\
\SetCell[r=3]{l} 2
  & TGTU (\textdegree C) & 5.51 & 0.92 & 28.13 & 0.15 & 0.23 \\
  & RUL (cycles)         & 458  & 0.99 & 2274  & 0.73 & 0.93 \\
  & DTGT (\textdegree C) & 5.73 & 0.93 & 27.79 & 2.02 & 1.78 \\
\SetCell[r=3]{l} 3
  & TGTU (\textdegree C) & 7.79 & 0.87 & 29.65 & 0.17 & 0.28 \\
  & RUL (cycles)         & 705  & 0.97 & 2473  & 0.87 & 1.18 \\
  & DTGT (\textdegree C) & 6.00 & 0.96 & 32.14 & 1.55 & 1.20 \\
\SetCell[r=3]{l} 4
  & TGTU (\textdegree C) & 6.58 & 0.97 & 31.02 & 0.16 & 0.20 \\
  & RUL (cycles)         & 654  & 0.96 & 2329  & 0.78 & 1.09 \\
  & DTGT (\textdegree C) & 5.92 & 0.98 & 35.31 & 2.30 & 1.70 \\
\SetCell[r=3]{l} 5
  & TGTU (\textdegree C) & 4.95 & 0.91 & 29.64 & 0.17 & 0.23 \\
  & RUL (cycles)         & 413  & 0.96 & 1718  & 1.20 & 2.15 \\
  & DTGT (\textdegree C) & 7.40 & 0.96 & 35.25 & 4.38 & 3.15 \\
\SetCell[r=3]{l} 6
  & TGTU (\textdegree C) & 10.14 & 0.71 & 27.30 & 0.14 & \num{2.27e4} \\
  & RUL (cycles)         & 464   & 0.96 & 1663  & 0.46 & 0.35 \\
  & DTGT (\textdegree C) & 5.93  & 0.95 & 30.24 & 1.09 & 2.99 \\
\bottomrule
\end{tblr}
\end{table*}

\FloatBarrier

\Cref{ehm:tbl:tgtu_by_engine_phase} isolates TGTU performance per engine and phase. Across engine--phase pairs, MAE spans from $2.22\,^{\circ}\mathrm{C}$ (engine~E, climb) to $14.77\,^{\circ}\mathrm{C}$ (engine~H, takeoff), and PICP ranges from $0.36$ (engine~H, climb) to $1.00$. This variability suggests that uncertainty calibration is not uniform across the fleet and that a subset of engines dominates the low-coverage behavior.

\Cref{ehm:tbl:rul_by_engine_phase} reports analogous RUL metrics per engine and phase. MAE ranges from $102$ cycles (engine~H, climb) to $1695$ cycles (engine~F, cruise), and while coverage is near unity in many cases, some scenarios such as engine~S in cruise show reduced PICP ($0.78$) with a large MED-CWC penalty. These per-engine results are useful for diagnosing where long-horizon uncertainty becomes overly conservative or poorly calibrated.

\Cref{ehm:tbl:tgtu_by_engine_phase_segment} provides the most granular view retained in the dissertation chapter, decomposing TGTU performance by engine, phase, and segment. This table helps localize which specific segment--phase combinations drive the phase- and engine-level aggregates, making it suitable for root-cause analysis and for designing targeted model improvements (e.g., segment-aware augmentation or calibration).

\Cref{ehm:tbl:by_esn_phase} provides a compact cross-task breakdown by engine serial number, flight phase, and task, enabling identification of outlier engines or regimes without pushing the quantitative discussion outside the main Results section. The wide spread of MAE, PICP, and MED-CWC across engines highlights that aggregate metrics can mask difficult cases; these tables therefore help prioritize targeted remediation (e.g., additional data, domain adaptation, or per-engine recalibration) for the worst-performing combinations.

Because these engine-level diagnostics are part of the substantive quantitative findings, they are kept in the main Results section rather than deferred to an appendix-style placement. The selected rows emphasize best-case behavior, worst-case behavior, and representative low-coverage regimes that drive the calibration trends discussed above.

\begin{table*}[!t]
\centering
\caption{Representative TGTU prediction metrics by engine and flight phase. The selected rows highlight the lowest-error case, the largest-error case, the lowest-coverage case, a representative low-coverage takeoff regime, and a wide-interval cruise case.}
\label{ehm:tbl:tgtu_by_engine_phase}
\small
\begin{tblr}{width=\textwidth,colspec={l l r r r r r},colsep=3pt}
\toprule
Engine & Phase & MAE & PICP & MPIW & NMPIW & CWC \\
\midrule
E & Climb & 2.22 & 1.00 & 23.40 & 0.24 & 0.24 \\
H & Takeoff & 14.77 & 0.45 & 28.99 & 0.10 & \num{8.37e9} \\
H & Climb & 11.76 & 0.36 & 22.09 & 0.20 & \num{1.29e12} \\
B & Takeoff & 8.95 & 0.70 & 25.85 & 0.11 & \num{2.87e11} \\
F & Cruise & 6.88 & 0.98 & 36.96 & 0.65 & 0.65 \\
\bottomrule
\end{tblr}
\end{table*}

\begin{table*}[!t]
\centering
\caption{Representative RUL prediction metrics by engine and flight phase. The selected rows highlight the smallest and largest MAE values and several cruise regimes with reduced empirical coverage.}
\label{ehm:tbl:rul_by_engine_phase}
\small
\begin{tblr}{width=\textwidth,colspec={l l r r r r r},colsep=3pt}
\toprule
Engine & Phase & MAE & PICP & MPIW & NMPIW & CWC \\
\midrule
H & Climb & 102 & 1.00 & 1364 & 1.17 & 1.17 \\
F & Cruise & 1695 & 1.00 & \num{2.96e4} & 443.48 & 443.48 \\
S & Cruise & 862 & 0.78 & 2246 & 1.76 & \num{2.53e5} \\
C & Cruise & 797 & 0.84 & 2177 & 0.87 & 163.63 \\
B & Cruise & 816 & 0.88 & 2267 & 0.89 & 36.13 \\
\bottomrule
\end{tblr}
\end{table*}

\begin{table*}[!t]
\centering
\caption{Representative TGTU metrics by engine, flight phase, and segment. The selected rows correspond to the lowest-coverage engine--phase--segment combinations in the expanded evaluation set and therefore localize the strongest calibration failures.}
\label{ehm:tbl:tgtu_by_engine_phase_segment}
\small
\begin{tblr}{width=\textwidth,colspec={l l r r r r r r},colsep=3pt}
\toprule
Engine & Phase & Segment & MAE & PICP & MPIW & NMPIW & CWC \\
\midrule
N & Climb & 1.00 & 15.98 & 0.04 & 21.02 & 0.25 & \num{1.53e19} \\
K & Climb & 1.00 & 17.22 & 0.06 & 21.72 & 0.27 & \num{6.58e18} \\
P & Takeoff & 1.00 & 19.18 & 0.10 & 23.99 & 0.09 & \num{2.18e17} \\
C & Climb & 1.00 & 14.25 & 0.11 & 21.49 & 0.19 & \num{2.73e17} \\
I & Takeoff & 2.00 & 17.04 & 0.13 & 24.65 & 0.10 & \num{7.47e16} \\
K & Takeoff & 1.00 & 20.42 & 0.13 & 25.29 & 0.12 & \num{7.62e16} \\
\bottomrule
\end{tblr}
\end{table*}

\begin{table*}[!t]
\centering
\caption{Representative engine serial number, phase, and task combinations drawn from the expanded evaluation set. These rows illustrate how accuracy and calibration vary across tasks for specific engines and operating regimes.}
\label{ehm:tbl:by_esn_phase}
\small
\begin{tblr}{width=\textwidth,colspec={l l l r r r r r},colsep=3pt}
\toprule
Engine & Phase & Task & MAE & PICP & MPIW & NMPIW & CWC \\
\midrule
E & Climb & TGTU (\textdegree C) & 2.22 & 1.00 & 23.40 & 0.24 & 0.24 \\
H & Takeoff & TGTU (\textdegree C) & 14.77 & 0.45 & 28.99 & 0.10 & \num{8.37e9} \\
H & Climb & RUL (cycles) & 102 & 1.00 & 1364 & 1.17 & 1.17 \\
S & Cruise & RUL (cycles) & 862 & 0.78 & 2246 & 1.76 & \num{2.53e5} \\
H & Takeoff & DTGT (\textdegree C) & 14.66 & 0.35 & 27.11 & 1.35 & \num{1.16e13} \\
F & Cruise & RUL (cycles) & 1695 & 1.00 & \num{2.96e4} & 443.48 & 443.48 \\
\bottomrule
\end{tblr}
\end{table*}

\FloatBarrier

\Cref{ehm:fig:tgt-timeseries} shows a representative time-series prediction for turbine gas temperature (TGTU) during the climb phase for engine \texttt{B} (segment~4). The mean estimate is shown alongside the Engine Health Management (EHM) reference while the prediction interval (PI) summarizes uncertainty at each time step; regimes where the PI widens indicate conditions where the model is less certain and where additional data or calibration could be beneficial.

\Cref{ehm:fig:tgtu-timeseries-S-takeoff} provides a complementary TGTU example in takeoff for engine \texttt{S}, where the thermal dynamics are more transient. This qualitative case illustrates how the uncertainty band adapts during rapid changes in temperature, supporting downstream use-cases that must balance responsiveness with robustness to measurement noise.

\Cref{ehm:fig:dtgt-timeseries-B-cruise} visualizes predictions for DTGT in cruise for engine \texttt{B}. Because DTGT emphasizes deviations from nominal behavior, the overlay of the predicted and EHM DTGT trajectories---together with the associated PI---provides a qualitative check that the model can follow residual trends while explicitly quantifying confidence around those deviations.

\Cref{ehm:fig:rul-timeseries-B} reports remaining useful life (RUL) estimates for engine \texttt{B} over a representative segment. In addition to the point prediction, the PI conveys the range of plausible RUL values at each cycle, which is critical for translating model output into maintenance actions (e.g., conservative planning based on lower quantiles).

\Cref{ehm:fig:rul-timeseries-O-takeoff} shows RUL performance on a different engine and operating regime (takeoff for \texttt{O}). Presenting multiple engines/phases helps assess whether uncertainty behavior is consistent across conditions and highlights any regime-specific degradation in accuracy or calibration.

\Cref{ehm:fig:dtgt-timeseries-S-takeoff} provides a takeoff DTGT example for engine \texttt{S}. The predicted DTGT trajectory and PI offer a compact representation of both deviation magnitude and confidence, which is useful when DTGT is later fed into threshold-based degradation logic or alerting.

\Cref{ehm:fig:rul-pdf} illustrates the full probabilistic output for RUL in the form of a predicted probability distribution. This distributional view goes beyond a single mean value by exposing tail risk (probability of short remaining life) and enables computing prediction intervals directly from quantiles, supporting risk-aware maintenance scheduling.

%%%%%%%%%%%%%%%%%%%%%%%%%%%%%%%%%%%%%%%%%%%%%%%%%%%%%%%%%%%%%%%%%%%%%%%%%%%%%%%%

\FloatBarrier
\section{Conclusions}
\label{ehm:sec:Conclusions}

This paper presents a scientific machine learning framework for turbine prognostics that couples
high-fidelity sequence modeling with uncertainty quantification (UQ) to support robust decision-making.
The proposed approach jointly models turbine gas temperature untrimmed (TGTU), remaining useful life (RUL),
and degradation-related quantities while producing prediction intervals (PIs) that can be evaluated
and monitored using established PI metrics (e.g., PICP, MPIW, NMPIW, and CWC) \cite{Khosravi2011Review}.

\paragraph{Architecture summary}
The end-to-end pipeline begins with preprocessing and sequence construction (e.g., scaling/detrending and
a sliding-window representation). A shared temporal encoder then maps each sensor window into a latent
representation using a residual bidirectional LSTM stack followed by an attention pooling mechanism.
Task-specific prediction heads are attached to the shared representation. In particular, the primary
prognostic targets are modeled using Mean--Variance Estimation (MVE), where each head outputs both a
predictive mean and a predictive variance, enabling Gaussian predictive distributions and closed-form
prediction intervals. Additional heads can be used to predict degradation attributes (e.g., performance
level, degradation rate, and time-to-threshold) within the same multi-task framework. Each encoder component addresses a specific modeling requirement: convolutional filtering/downsampling for local patterns, bidirectional recurrence for longer temporal context, and attention pooling for focusing on the most informative time steps. This modular design offers a practical compromise between simpler single-task encoders and more complex sequence models when working with heterogeneous real fleet data.

\paragraph{Novel aspects}
Several aspects distinguish this work from more conventional prognostic pipelines:
(i) a unified, multi-objective network simultaneously predicts multiple health indicators (TGT, RUL,
and degradation attributes) from shared representations;
(ii) task trade-offs are handled in a principled way via learned task weighting, which reduces the need
for manual loss-tuning in multi-task learning \cite{KendallGal2018};
(iii) uncertainty is treated as a first-class output by design (via MVE), and PI quality is assessed using
rigorous PI metrics such as CWC \cite{Khosravi2011Review};
(iv) the evaluation is stratified by operational context (flight phase and maintenance segment),
enabling data-centric interpretation beyond aggregate test-set scores; and
(v) the methodology is trained and evaluated on real fleet data (not only synthetic benchmarks), which is
important because field data include non-stationarity, maintenance resets, and regime shifts that can strongly
affect both predictive accuracy and uncertainty calibration.

%%%%%%%%%%%%%%%%%%%%%%%%%%%%%%%%%%%%%%%%%%%%%%%%%%%%%%%%%%%%%%%%%%%%%%

\paragraph{Performance overview}
Across the evaluated dataset, the framework achieves low predictive error for TGTU and RUL while
maintaining high PI coverage and reasonable PI width. Phase- and segment-wise analyses show that the
model generalizes across operating regimes, with performance varying in interpretable ways as the engine
progresses through different phases/segments. Notably, PI coverage remains strong overall, suggesting the
learned predictive distributions remain reasonably calibrated in aggregate, with under-coverage in specific regimes.

\paragraph{Observations}
While aggregate metrics indicate strong performance, the stratified analysis reveals that the \emph{median}
CWC can differ substantially for specific combinations of engine ID, flight phase, and flight segment.
Because CWC jointly captures both coverage behavior and interval width \cite{Khosravi2011Review}, a pronounced
shift in median CWC relative to the rest of the dataset suggests those subsets may be \emph{abnormal} in the
sense of being inherently different (e.g., a distinct operating regime, sensor characteristics, or
sensor collection records). Given how well the network models the majority of the data, such
discrepancies plausibly reflect a real distributional difference localized to those data files containing the given combination of engine ID, flight phase, and flight segment. Importantly,
this does \emph{not} necessarily imply the data are faulty; however, it is a strong quantitative signal that
the certain engine ID/phase/segment data warrants deeper investigation. This observation naturally
motivates the future-work direction of systematic abnormality detection.

\paragraph{Future work}
Two follow-on directions are especially promising:

\begin{enumerate}
  \item An applied survey of UQ methods for turbine gas temperature prediction.
  While this paper emphasizes MVE-based probabilistic regression, future work should benchmark additional
  UQ approaches for TGT/DTGT in a controlled, applied survey. Candidates include the major PI construction
  families reviewed in \cite{Khosravi2011Review} (e.g., quantile-regression and ensemble-based intervals),
  distribution-free conformal prediction and related conformalized constructions \cite{AngelopoulosBates2021Conformal},
  as well as modern deep-learning UQ baselines such as Monte Carlo dropout \cite{GalGhahramani2016} and deep ensembles
  \cite{Efron1979,lakshminarayanan2017deep}. In addition to aggregate metrics, the comparison should explicitly
  evaluate calibration degradation under regime shift \cite{Ovadia2019CanYouTrust} and report the accuracy--cost trade-offs
  that matter for deployment on turbine datasets.

  \item An abnormality detection framework using CWC discrepancy monitoring.
  The discovered large discrepancies in median CWC motivate a dedicated abnormality detection framework.
  A practical extension would track PI-quality statistics (e.g., median CWC and related PI metrics) by engine ID, flight phase, flight
  segment, and trigger alerts when deviations exceed robust thresholds. This can be connected to
  established anomaly detection ideas \cite{chandola2009survey} as well as uncertainty- or out-of-distribution-aware monitoring baselines  \cite{hendrycks2017baseline}, yielding a lightweight notification layer that flags subsets with unusually large
  PI degradation for targeted data and system review.
\end{enumerate}

%%%%%%%%%%%%%%%%%%%%%%%%%%%%%%%%%%%%%%%%%%%%%%%%%%%%%%%%%%%%%%%%%%%%%%
\section{Funding}
This research was sponsored by Rolls-Royce and Virginia Tech.

%%%%%%%%%%%%%%%%%%%%%%%%%%%%%%%%%%%%%%%%%%%%%%%%%%%%%%%%%%%%%%%%%%%%%%
\section{Data Availability}

The data supporting the findings of this study were made available from Rolls-Royce plc, but restrictions apply to the availability of these data, which were used under license for the current study and are not publicly available. The point of contact for data requests is changminson@vt.edu.

%%%%%%%%%%%%%%%%%%%%%%%%%%%%%%%%%%%%%%%%%%%%%%%%%%%%%%%%%%%%%%%%%%%%%%
\section{Acknowledgments}

The authors would like to express sincere appreciation to Rolls-Royce for sharing Engine Health Management (EHM) data and technical insights. The authors extend special thanks to Jin-Sol Jung for technical discussion about mechanical engineering aspects pertaining to the EHM data.

%%%%%%%%%%%%%%%%%%%%%%%%%%%%%%%%%%%%%%%%%%%%%%%%%%

\section{Conflict of Interest}
Authors Gavan Burke, Rekha Sundararajan, Andrew Rimell, and Gregory Steinrock were employed by Rolls-Royce. The remaining authors declare that the research was conducted in the absence of any commercial or financial relationships that could be construed as a potential conflict of interest.

%%%%%%%%%%%%%%%%%%%%%%%%%%%%%%%%%%%%%%%%%%%%%%%%%%
\section{Declaration of Generative AI}
During the preparation of this work, the authors used OpenAI's ChatGPT in order to assist with crafting the formal notation of the presented methodologies and their underlying concepts. Furthermore, the tool was used to enhance coding, copy editing, and literature research. After using this tool/service, the authors reviewed and edited the content as needed and take full responsibility for the content of the published article.

\bibliographystyle{IEEEtran}
\bibliography{references}

\end{document}